\documentclass[12pt]{article}
\usepackage{latexsym,epsfig,graphicx,amsmath,amssymb,amscd,undertilde,multirow,chicago,psfrag,paralist,dsfont}
\usepackage[titletoc]{appendix}

\usepackage{subcaption}

\usepackage{hyperref}

\usepackage{epstopdf}
\usepackage[american]{babel}
\usepackage{xr}

\newcommand{\thetavec}{{\boldsymbol{\theta}}}

\newcommand{\Sigmavec}{{\boldsymbol{\Sigma}}}

\newcommand{\muvec}{{\boldsymbol{\mu}}}

\newcommand{\zvec}{{\boldsymbol{z}}}

\newcommand{\Ivec}{{\boldsymbol{I}}}
\newcommand{\Indfun}{{\mathds{1}}}

\newcommand{\E}{\mathds{E}}

\newcommand{\pr}{{\rm Pr}}

\newcommand{\NOR}{{\rm N}}

\newcommand{\Sigmahat}{\widehat{\Sigmavec}}

\newcommand{\muvechat}{\widehat{\muvec}}
\newcommand{\tran}{^\top}
\newcommand{\Xb}{{\boldsymbol{X}}}
\newcommand{\fca}{{\text{(1)}}}
\newcommand{\fcb}{{\text{(2)}}}
\newcommand{\fcc}{{\text{(3)}}}
\newcommand{\fcd}{{\text{(4)}}}
\newcommand{\Tset}{{\mathsf{T}}}
\newcommand{\Sset}{{\mathsf{S}}}
\newcommand{\Cset}{{\mathsf{C}}}
\newcommand{\Oset}{{\mathsf{O}}}
\DeclareMathOperator*{\argmin}{argmin}

\begin{document}
	
	%%%%%%%%%%%%TITLE%%%%%%%%%%%%%%%%%%%%%%%%%%%%%%%%%%%
	\title{Deep Neural Network Identification of \textit{Limnonectes} Species and New Class Detection Using Image Data}
	%%%%%%%%%%%%%%%%%%%%%%%%%%%%%%%%%%%%%%%%%%%%%%%%%%%%%%%%%%%%%%%%%%%%%%%%%%%%%%%%%%%%%%%%%%%%%%%%%%%%%%%%%%%%%%%%%

	\author{Li Xu$^1$, Yili Hong$^2$, Eric P. Smith$^2$, David S. McLeod$^{3,4}$,\\
		Xinwei Deng$^2$, and Laura J. Freeman$^2$\\[1.5ex]
		{\small $^1$Department of Biostatistics, Harvard University, Cambridge, MA 02138}\\
		{\small $^2$Department of Statistics, Virginia Tech, Blacksburg, VA 24061}\\
		{\small $^3$Murphy Deming College of Health Sciences, Mary Baldwin University, Staunton, VA 24401}\\
        {\small $^4$Department of Vertebrate Zoology, National Museum of Natural History,}\\[-1ex]
        {\small Smithsonian Institution, Washington, DC 20560}
	}

	\date{}

	\maketitle
	%%%%%%%%%%%%%%%%%%%%%%%%%%%%%%%%%%%%%%%%%%%%%%%%%%%%%%%%%%%%%%%%%%%%%%%%%%%%%%%%%%%%%%%%%%%%%%%%%%%%%%%%%%%%%%%%
	\begin{abstract}

As is true of many complex tasks, the work of discovering, describing, and understanding the diversity of life on Earth (viz., biological systematics and taxonomy) requires many tools.  Some of this work can be accomplished as it has been done in the past, but some aspects present us with challenges which traditional knowledge and tools cannot adequately resolve.  One such challenge is presented by species complexes in which the morphological similarities among the group members make it difficult to reliably identify known species and detect new ones. We address this challenge by developing new tools using the principles of machine learning to resolve two specific questions related to species complexes. The first question is formulated as a classification problem in statistics and machine learning and the second question is an out-of-distribution (OOD) detection problem. We apply these tools to a species complex comprising Southeast Asian stream frogs (\emph{Limnonectes kuhlii} complex) and employ a morphological character (hind limb skin texture) traditionally treated qualitatively in a quantitative and objective manner. We demonstrate that deep neural networks can successfully automate the classification of an image into a known species group for which it has been trained. We further demonstrate that the algorithm can successfully classify an image into a new class if the image does not belong to the existing classes. Additionally, we use the larger MNIST dataset to test the performance of our OOD detection algorithm. We finish our paper with some concluding remarks regarding the application of these methods to species complexes and our efforts to document true biodiversity. This paper has online supplementary materials.


\end{abstract}

	%%%%%%%%%%%%%%%%%%%%%%%%%%%%%%%%%%%%%%%%%%%%%%%%%%%%%%%%%%%%%%%%%%%%%%%%%%%%%%%%%%%%%%%%%%%%%%%%%%%%%%%%%%%%%%%%%%%%%
	\section{Introduction}
	%%%%%%%%%%%%%%%%%%%%%%%%%%%%%%%%%%%%%%%%%%%%%%%%%%%%%%%%%%%%%%%%%%%%%%%%%%%%%%%%%%%%%%%%%%%%%%%%%%%%%%%%%%%%%%%%%%%%%
\subsection{Background and Motivation}

We are facing a biodiversity crisis. At present we have discovered, identified, and described only a very small fraction of the organisms with which we share this planet (e.g., \citeNP{Sweetlove2011}). Unfortunately, because of habitat loss, climate change, and other anthropogenic influences, it is likely that we will lose this diversity faster than we can recognize it (e.g., \shortciteNP{Cowieetal2022}). Our ability to accurately identify and delineate biological diversity is critically important because it influences our understanding of true biological richness and informs conservation efforts to safeguard these natural resources. Two critical tasks involved in understanding biological diversity are (1) the ability to correctly identify an individual organism to the taxonomic rank of species, and (2) the ability to detect individuals that may represents new species that are undescribed and therefore unrecognized by the scientific community. Traditionally, we have relied on the expertise of taxonomic specialists who drew from an understanding of morphology and natural history to identify new species. In recent decades, tools such as DNA sequencing and phylogenetic analyses have been employed by systematists in this pursuit. These new methods have advanced our rate of species discovery (e.g., \shortciteNP{YAMASAKI2017177}) and revealed species complexes in which morphological similarity has obfuscated true diversity (e.g., \shortciteNP{Freudensteinetal2016}).

We consider a species complex to be a group of two or more phylogenetically related species-level taxa that share such morphological similarity that group members cannot easily be distinguished from one another (\shortciteNP{Scherzetal2019} and \shortciteNP{deSousa-Paulaetal2021}). Examples of species complexes can be found across all domains of life and present unique challenges to taxonomists and conservation biologists alike who would benefit from a new suite of tools that facilitate the recognition and detection of both known and unknown species.

\begin{figure}
		\begin{center}
			\begin{tabular}{cc}
				\includegraphics[width=0.45\textwidth]{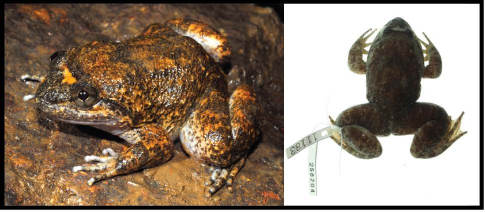} &
				\includegraphics[width=0.45\textwidth]{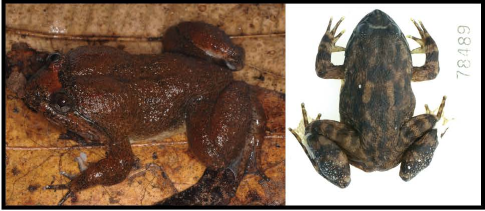} \\[-1ex]
				(a) Clade 4   & (b) Clade 5  \\[1ex]
				\includegraphics[width=0.45\textwidth]{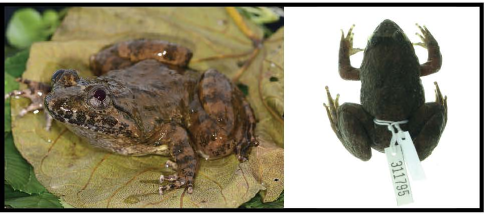} &
				\includegraphics[width=0.45\textwidth]{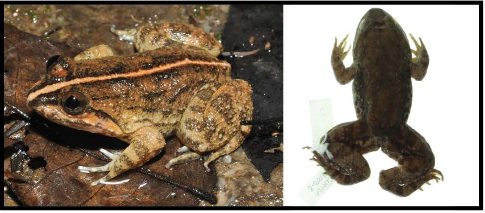} \\[-1ex]
				(c) Clade 8   & (d) Clade 12  \\[1ex]
			\end{tabular}
        \caption{Living and preserved examples of the four species (clades) from the \emph{L. kuhlii} complex sampled in this study. Following McLeod (2010) species are represented by clade numbers: \emph{L. fastigatus} (Clade 4), \emph{L. kiziriani} (Clade 5), \emph{L. bannaensis} (Clade 8), and \emph{L. taylori} (Clade 12). Photos in life by Nguyen Thanh Luan (\emph{L. bannaensis} and \emph{L. kiziriani}), Jodi Rowley (\emph{L. fastigatus}), and Attapol Rujirawan (\emph{L. taylori}), all other photos by DSM.}
			\label{fig:example.frog.images}
		\end{center}
	\end{figure}

In this paper, we focus on one such species complex comprising a group of Southeast Asian stream frogs. The frogs allied to the \emph{Limnonectes kuhlii} (abbreviated as \emph{L. kuhlii}) complex were considered to be a single species for about 200 years. Broadly distributed throughout Southeast Asia these frogs share a tremendous amount of morphological and ecological similarities. On the basis of DNA evidence alone, molecular phylogenetic studies (e.g., \shortciteNP{Yeetal2007}, \citeNP{McLeod2008}, \citeNP{MCLEOD2010}, \shortciteNP{Matsuietal2010}, and \shortciteNP{Stuartetal2020}) recovered more than 22 unique evolutionary lineages representing potentially new species within the \emph{L. kuhlii} complex. Using these data in combination with morphological evidence, subsequent taxonomic studies have described several new species from within the \emph{L. kuhlii} complex (e.g., \emph{L. megastomias} in \citeNP{McLeod2008}, \emph{L. nguyenorum} in \shortciteNP{McLeodetal2012}, \emph{L. isanensis} in \shortciteNP{Suwannapoometal2017}, and \emph{L. fastigatus} in \shortciteNP{Stuartetal2020}). Figure~\ref{fig:example.frog.images} shows four examples of frogs from four different clades (i.e., lineages).

\begin{figure}
		\begin{center}
			\begin{tabular}{cccc}
				\includegraphics[width=0.23\textwidth]{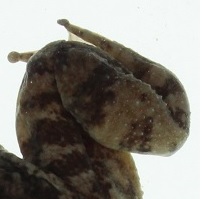} &
				\includegraphics[width=0.23\textwidth]{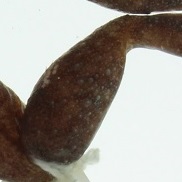} &
				\includegraphics[width=0.23\textwidth]{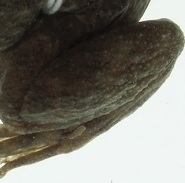} &
				\includegraphics[width=0.23\textwidth]{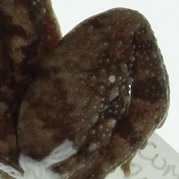} \\%[-1ex]
				 (a) Clade 4   & (b) Clade 5  & (c) Clade 8   & (d) Clade 12  \\[1ex]
			\end{tabular}
			
\caption{Example leg images from four clades of the \emph{L. kuhlii} species.}\label{fig:example.frog.leg.images}
		\end{center}
	\end{figure}

Despite these advances, delineating species boundaries and describing additional new species within this complex has been confounded by the high degree of morphological similarity among the genetically distinct lineages. One of the diagnostic morphological characters that has proven useful in distinguishing between species in the \emph{L. kuhlii} complex is the texture of the skin on the frog's hind legs. This texture (created by the patterning of raised tubercles) is traditionally described qualitatively on the spectra of ``smooth to rough'' or ``dense to sparse'', but without quantification of any kind. The pattern of tuberculation is consistent within a species but varies between species (\shortciteNP{McLeodetal2011}, \shortciteNP{McLeodetal2012}, and \shortciteNP{Stuartetal2020}). The consistent yet subjectively qualitative nature of this character opens the door for applying machine-learning-based methods to standardize and automate species recognition and the detection of new species. Thus our specific research questions in this paper are: 1) whether machine-learning methods can use a qualitative character, such as the pattern of tuberculation on frog legs, to classify individuals into known classes (viz., species), and 2) whether a machine-learning-based tool can be developed to detect new classes (viz., new species). Figure~\ref{fig:example.frog.leg.images} shows examples of the texture of the frog legs from the four clades for the four samples as shown in Figure~\ref{fig:example.frog.images}, which highlights visible differences in tubercle structure on the hind limbs.

\subsection{Research Questions}

The first research question can be formulated as an image classification problem. Thus, deep-learning techniques can be applied. Deep learning has proven to be a powerful tool in resolving image classification problems across diverse applications (e.g., \shortciteNP{Krizhevskyetal2012}, and \citeNP{Rawatetal2017}).  \shortciteN{Norouzzadehetal2018} applied a deep-learning method to monitor wildlife using camera-trap images, and \shortciteN{Wangetal2021} applied deep-learning algorithm on CT images to screen for COVID-19 disease. Whereas the application of machine learning is commonly used in species identification (e.g., iNaturalist, www.inaturalist.org, and PictureThis, www.picturethisai.com) and there has been some application of machine-learning and deep-learning methods to detect species from audio files of frog calls (e.g., \shortciteNP{Huangetal2009}, \shortciteNP{Hassanetal2017}, and \shortciteNP{Xieetal2017}), no study has applied these methods to problems associated with species complexes and new species detection. In this paper, we develop a convolutional neural network (CNN) model that can serve as an automatic identifier or classifier based on image/pattern recognition.  This tool works in such a way that a photograph of an individual frog can be classified into one of the existing species using an existing set of images.

The second research question can be formulated as an out-of-distribution detection (OOD) problem (e.g., \shortciteNP{hendrycks2017a}, and \shortciteNP{liang2018enhancing}). The developed new class detection (NCD) tool allows for an image to be ``rejected'', suggesting the detection of a potential new species. CNN and OOD techniques are well developed in the machine-learning discipline. For example, \shortciteN{Leeetal2018} presents a simple unified framework for detecting OOD samples under the setting of adversarial attacks to neural network. The applications of CNN and OOD tools to biological research in the context of species complexes are new, and our approach provides quantitative tools to biological scientists working in the area. Our NCD methods extend the framework in \shortciteN{Leeetal2018} and customize it for the frog NCD problem.

\subsection{Overview}
	
	The rest of the paper is organized as follows. Section~\ref{sec:data.description.preprocessing} describes the collection and preprocessing of the frog image data used in this study. Section~\ref{sec:method.development} describes the CNN model for classification and describes a new method for NCD. Section~\ref{sec:results} presents the classification and NCD results for the frog datasets and the NCD results for the MNIST dataset (\citeNP{Deng2012}) for further validation. Section~\ref{sec:conluding.remarks} contains some concluding remarks.

	%%%%%%%%%%%%%%%%%%%%%%%%%%%%%%%%%%%%%%%%%%%%%%%%%%%%%%%%%%%%%%%%%%%%%%%%%%%%%%%%%%%%%%%%%%%%%%%%%%%%%%%%%%%%%%%%%%%%%
\section{Description of Image Data and Preprocessing}\label{sec:data.description.preprocessing}
	%%%%%%%%%%%%%%%%%%%%%%%%%%%%%%%%%%%%%%%%%%%%%%%%%%%%%%%%%%%%%%%%%%%%%%%%%%%%%%%%%%%%%%%%%%%%%%%%%%%%%%%%%%%%%%%%%%%%%

In this paper, we use image data collected from a subset of lineages identified as part of the \emph{Limnonectes kuhlii} complex (\citeNP{MCLEOD2010}). We chose to build our neural network using samples representing lineages 4 (\emph{Limnonectes fastigatus}, \shortciteNP{Stuartetal2020}), 5 (\emph{Limnonectes kiziriani}, \shortciteNP{Phametal2018}), 8 (\emph{Limnonectes bannaensis}, \shortciteNP{Yeetal2007}), and 12 (\emph{Limnonectes taylori}, \shortciteNP{Matsuietal2010}) based on the relatively large number of sample images available. A cleaned subset of those images is used to develop and evaluate the classifier for the species.

\begin{figure}
		\begin{center}
			\begin{tabular}{cc}
				\includegraphics[width=.35\linewidth, height=.27\linewidth]{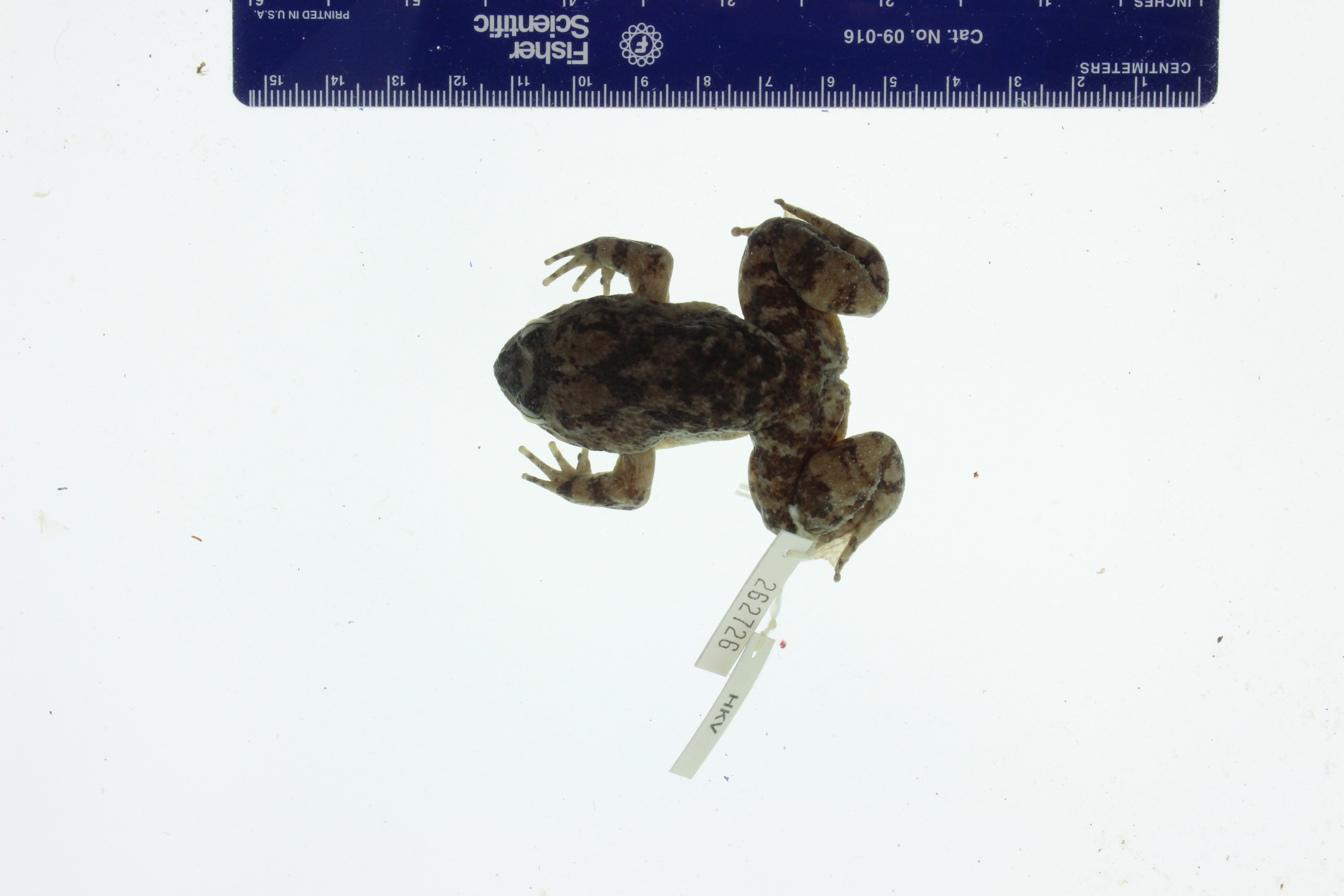}&
				\includegraphics[width=.35\linewidth, height=.27\linewidth]{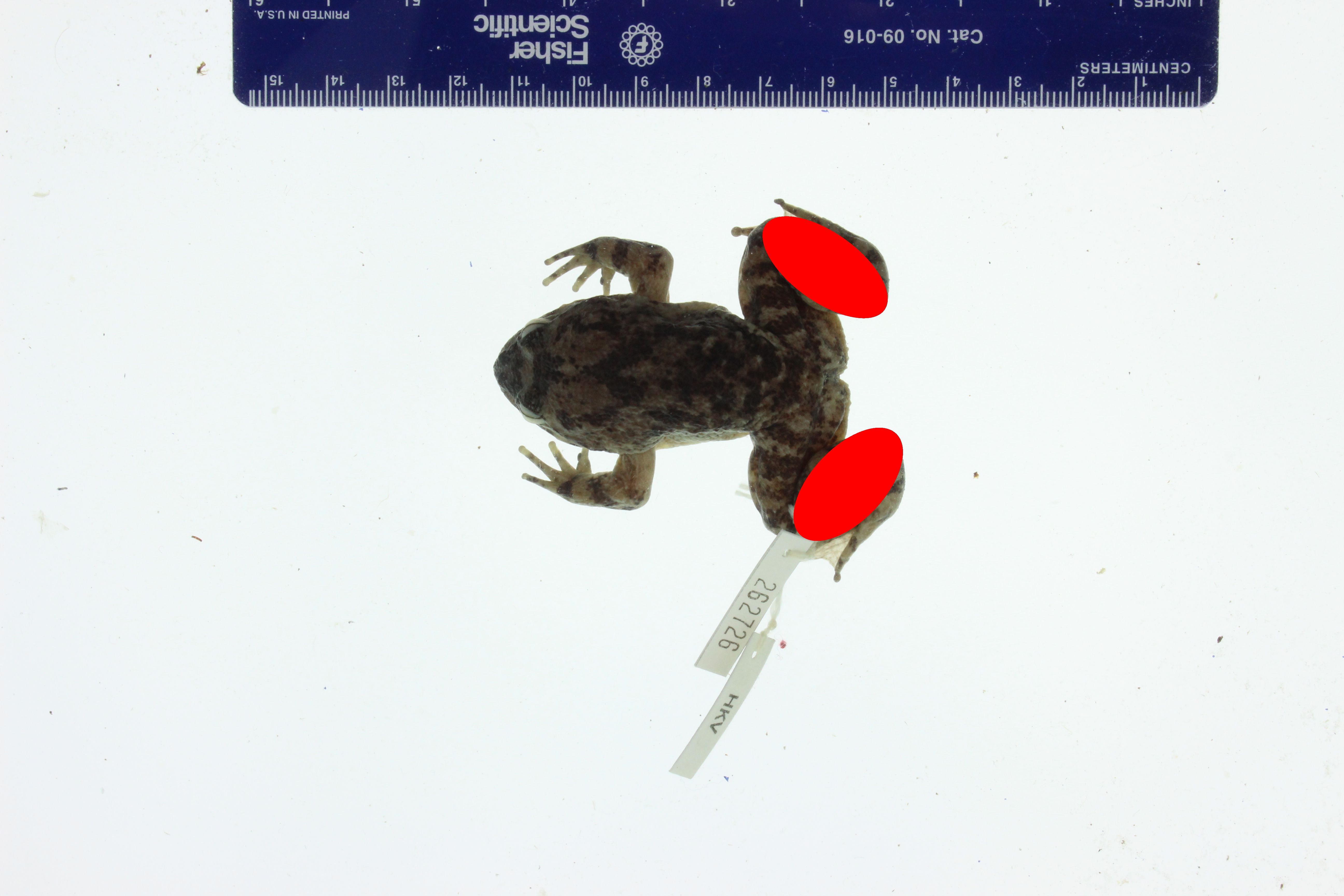}\\
				(a) Original Image & (b) Leg Mark\\[2ex]
				\includegraphics[width=.35\linewidth, height=.27\linewidth]{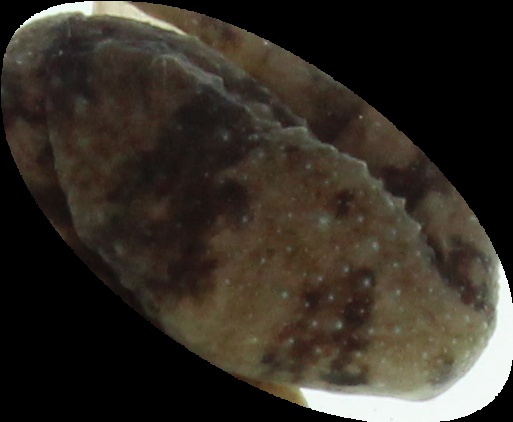}&
				\includegraphics[width=.35\linewidth, height=.27\linewidth]{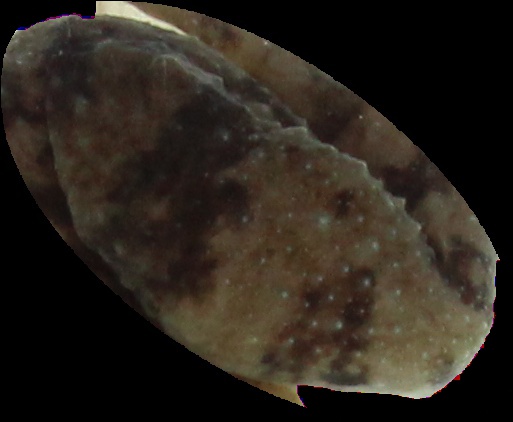}\\
				(c) Legs with Mixed Background& (d) Adjusted Black Background
			\end{tabular}
			\caption{Illustration of the preprocessing of the two frog legs.}\label{fig:leg_w}
		\end{center}
	\end{figure}

Our first question is whether a deep-learning algorithm applied to the texture (tuberculation) of the skin on the hind legs complex can successfully differentiate among species in the \emph{L. kuhlii} complex. We therefore focus on images of the legs of the frogs. In order to utilize these raw images, some data preprocessing is needed to ensure consistency among the images. We found that raw specimen photographs vary in clarity due to noise factors such as the inclusion of scale bars and specimen tags as shown in Figure~\ref{fig:leg_w}(a). In order to reduce image noise we cropped out the legs of the frog (indicated by two red ellipses in Figure~\ref{fig:leg_w}(b)). We first manually label the two
	frog legs with red ellipses. Then the red part is cropped as shown in Figure~\ref{fig:leg_w}(c). In addition, the white background is replaced by black as shown in Figure~\ref{fig:leg_w}(d).
	To increase the sample size, the left and right legs are treated as two different images
	after pre-processing. Those images with legs obstructed by noise items, for example, specimen tags, or non-specimen structures such as human fingers, were removed from the sample set. A summary of image counts is given in Table~\ref{tab:frog.counts}. In total, we have 193 images, which we consider to be a moderate sample size for implementing some deep-learning models.

	After the image pre-processing, we resize our images to be $512\times 512$ in color pixels. For each individual pixel, there are three channels for red, green, and blue color components, with numbers ranging from 0 to 255. For example, $(255, 255, 255)\tran$ represents a white pixel and $(0, 0, 0)\tran$ represents a black pixel. Thus, all the pixels in an image forms a 3-dimensional (3-D) tensor with size $3\times512\times512$.

Some notations are needed for later development. The (3-D) image tensor with size $3\times512\times512$ is denoted as $\Xb_i$, where $i$ is the index for image. For notation convenience, we use a class label ranging from 1 to 4 to represent the clades, as shown in Table~\ref{tab:frog.counts}. For example, class label 1 corresponds to clade 4. We let $y_i$ be the class label of the image and $y_i$ takes values in $\{j;\, j=1, \dots, m\}$, where $m$ is the number of label classes. Here, $m=4$ as shown in Table~\ref{tab:frog.counts}. The data are denoted by $\{y_i, \Xb_i\}$ and $i$ is the index for the sample. The total number of samples/images is 193, as shown in Table~\ref{tab:frog.counts}.

	\begin{table}
		\begin{center}
		\caption{The total number of leg images in each clade after image preprocessing. There are frog samples with legs covered by the label, making the number of images an odd number.}\label{tab:frog.counts}
		\begin{tabular}{cccccc}\hline\hline
			Clade & Clade 4 & Clade 5 & Clade 8 & Clade 12 &Total\\
			\hline
			Number of leg images		&		44	&	22		&	77		&	50	&193	\\\hline
Class label & 1 & 2 & 3 & 4\\\hline\hline
	    \end{tabular}
        \end{center}
\end{table}

%%%%%%%%%%%%%%%%%%%%%%%%%%%%%%%%%%%%%%%%%%%%%%%%%%%%%%%%%%%%%%%%%%%%%%%%%%%%%%%%%%%%%%%%%%%%%%%%%%%%%%%%%%%%%%%%%%%%%
\section{Methods Development}\label{sec:method.development}	
%%%%%%%%%%%%%%%%%%%%%%%%%%%%%%%%%%%%%%%%%%%%%%%%%%%%%%%%%%%%%%%%%%%%%%%%%%%%%%%%%%%%%%%%%%%%%%%%%%%%%%%%%%%%%%%%%%%%%

%%%%%%%%%%%%%%%%%%%%%%%%%%%%%%%%%%%%%%%%%%%%%%%%%%%%%%%%%%%%%%%%%%%%%%%%%%%%%%%%%%%%%%%%%%%%%%%%%%%%%%%%%%%%%%%%%%%%%
	\subsection{Convolutional Neural Network for Classification}\label{sec:CNN.classification}
	%%%%%%%%%%%%%%%%%%%%%%%%%%%%%%%%%%%%%%%%%%%%%%%%%%%%%%%%%%%%%%%%%%%%%%%%%%%%%%%%%%%%%%%%%%%%%%%%%%%%%%%%%%%%%%%%%%%%%
	
	In this section, we develop a classifier that can predict the species based on frog leg
	images. We use on deep neural networks (e.g., \citeNP{Goodfellowetal2016}). In
	particular, we use the convolutional neural networks (CNN) for the prediction that can
	provide classification based directly on the images of frog legs. In the literature, the CNN has been
	found to be a powerful tool in image classification. A CNN consists of an input layer, an
	output layer, and multiple hidden layers. The hidden layers consist of convolutional
	layers, pooling layers, and fully connected layers. Sometimes normalization layers and
	dropout layers are also used. More details can be found in Chapter 9 of
	\citeN{Goodfellowetal2016}.

\begin{figure}[ht]
		\begin{center}
		\includegraphics[width=0.9\textwidth]{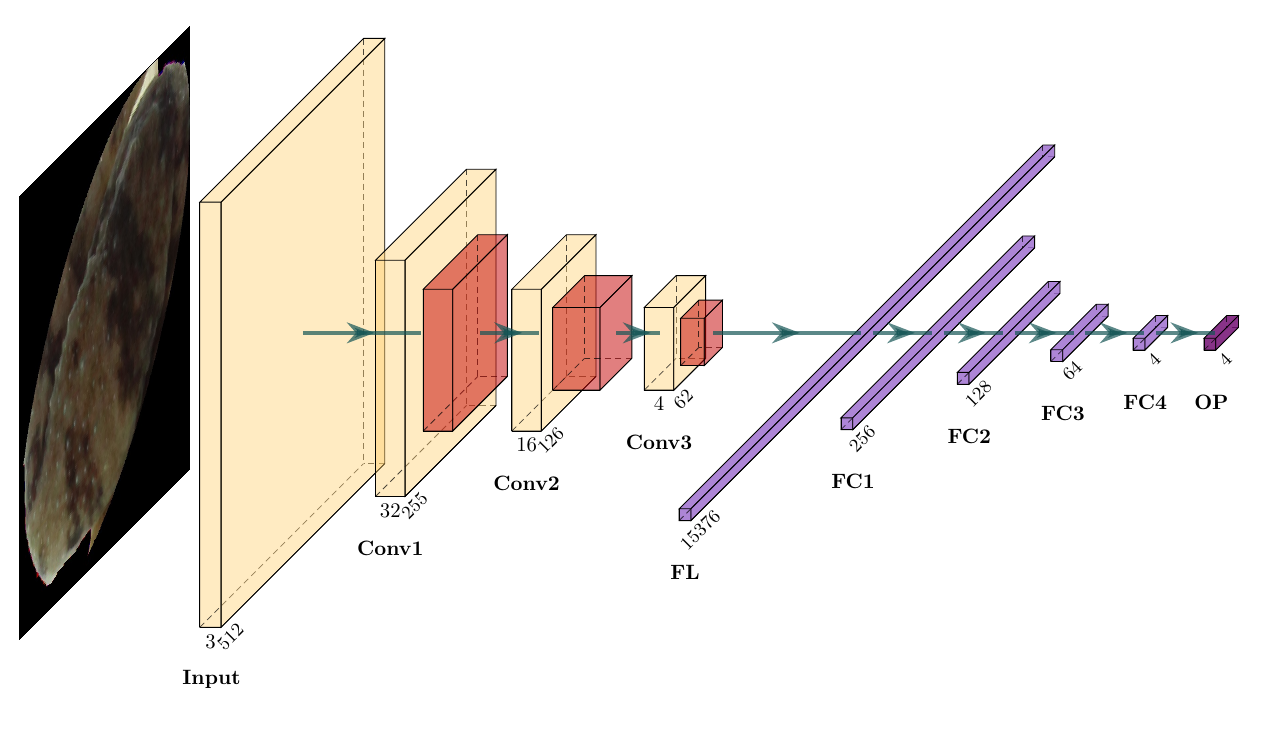}
		\caption{Diagram of CNN that is used for classification on the frog data. Here ``Conv1'' means convolution layer 1, ``MP" means maximum pooling, ``FL'' means flatten layer, ``FC1'' means fully connected layer 1, and ``OP'' means output layer.}\label{fig:FrogCNN}
\end{center}
	\end{figure}

Here we introduce the network architecture used for the frog images. Figure~\ref{fig:FrogCNN} illustrates the structure of the CNN used in this paper, which is a
	relatively simple structure due to the limited number of images, compared to the
	magnitude of tens of thousands of images in some other applications. We also tried CNN with more complicated structures. However, the prediction results are not getting better. For the CNN structure in Figure~\ref{fig:FrogCNN}, the input image is $\Xb_i$, which is a $(3,512,512)$ tensor. The
	kernel size for the convolution is $(3,3)$. After the convolution, we do maximum pooling. The kernel size for the maximum pooling is $(2,2)$ with a stride of $(2,2)$ to reduce the image size. Between each convolutional layer, batch normalization is applied. For
	each fully connected layer, the random dropout rate is set to 0.8.

As shown in Figure~\ref{fig:FrogCNN}, we have three convolutional layers after the input layer, which are labeled as Conv1, Conv2, and Conv3, respectively. The number of filters we use in Conv1, Conv2, and Conv3 are 32, 16, and 4, respectively. After application of Conv3, we obtain a $(4, 62, 62)$ tensor and it is flattened to a $15{,}376\times1$ vector. The flattened vector then goes through four fully connected layers, labeled by FC1, FC2, FC3, and FC4, respectively. The lengths of those four output vectors are 256, 128, 64, and 4, respectively. For convenience, we denote the outputs of those fully connected layers by
\begin{align}\label{eqn:notation.output.layer}
\zvec_{i}^{\fca}, \zvec_{i}^{\fcb}, \zvec_{i}^{\fcc}, \text{ and } \zvec_{i}^{\fcd},
\end{align}
respectively. Here, $i$ is the index for the input image. In the output layer, as labeled ``OP'' in Figure~\ref{fig:FrogCNN}, we use the softmax function to map $\zvec_{i}^{\fcd}$ to a probability. That is
\begin{align}
p_{ij}= \frac{\exp(z_{ij}^{\fcd})}{\sum_{k=1}^m \exp(z_{ik}^{\fcd})}, \quad j=1,\dots,m,
\end{align}
where $\zvec_{i}^{\fcd}=(z_{i1}^{\fcd}, \dots, z_{im}^{\fcd})\tran$. Here, $j$ is the index for class label.

Let $\Sset$ be the set of indexes for samples used in model training and let $\Tset$ be the set of indexes for samples used in testing. Note that we do not have a validation set because of the small sample size and the fact that we do not have hyper-parameters in the model training. Let $\thetavec$ be the set of all parameters in the CNN model. The categorical entropy is used as the loss function, that is,
\begin{align}
L(\thetavec)=-\sum_{i\in \Sset}\delta_{ij}\log(p_{ij}),
\end{align}
where $\delta_{ij}=\Indfun(y_i=j)$ and $\Indfun(\cdot)$ is the indicator function. PyTorch (\shortciteNP{PyTorch}) is used for training the model to find the $\thetavec$ that minimizes the loss function.

	To measure the model prediction accuracy, we use $K$-fold cross validation. We have an unbalanced classification problem. If we split the data randomly across all four species, the training set may not have all four species. To ensure that all four species
	are in our training set and the testing set consists of equal proportions from each of the four species for each fold, we use a balanced cross validation. For each species with class label $j$, we denote the index set by $\Cset_j, j=1, \dots, m$. The process for balanced cross validation is described as follows.

\begin{inparaenum}
\item We randomly split $\Cset_j$ in to $K$ equal folds, which are denoted by $\Cset_{jk}, k=1, \dots, K$.

\item For a given $k$, construct the training set as $\Sset=\cup_{j}\cup_{l\neq k}\Cset_{jl}$ and the testing set is as $\Tset=\cup_{j}\Cset_{jk}$. Then train the model based on set $\Sset$ and make predictions for samples in set $\Tset$.

\item Repeat Step~2 for $k=1, \dots, K$.

\item Repeat Steps~1 to 3 to obtain different random splits.
\end{inparaenum}

By using a balanced cross validation, we can ensure that each class always appears in both the training and testing sets.

%%%%%%%%%%%%%%%%%%%%%%%%%%%%%%%%%%%%%%%%%%%%%%%%%%%%%%%%%%%%%%%%%%%%%%%%%%%%%%%%%%%%%%%%%%%%%%%%%%%%%%%%%%%%%%%%%%%%%
	\subsection{New Class Detection}\label{sec:NCD}
	%%%%%%%%%%%%%%%%%%%%%%%%%%%%%%%%%%%%%%%%%%%%%%%%%%%%%%%%%%%%%%%%%%%%%%%%%%%%%%%%%%%%%%%%%%%%%%%%%%%%%%%%%%%%%%%%%%%%%
	In this section, we describe our algorithms for new class detection (NCD) for frog species. To evaluate the performance of our NCD algorithm, we take one of the classes in frog clades, say class $k$, as the samples from the out-of-distribution (OOD) population. Then the rest of the three classes are used to build the model.

For a trained deep-learning model, the outputs of the intermediate layers, such as those in \eqref{eqn:notation.output.layer}, represent the extracted features from the sample. If a new sample does not belong to one of existing classes in the training set, then one would expect those extracted features would be different from those features from samples from existing classes. Thus, detecting new classes based on the outputs of intermediate layers can be treated as detecting outliers based on these multivariate outputs. Similar ideas have been used in the literature (e.g., \shortciteNP{Leeetal2018}). In this paper, we customize this idea for the our frog species detection problem and use parametric discriminant analyses for NCD.

Suppose we already have a trained deep neural network classification model, such as the one in Figure~\ref{fig:FrogCNN} (but with 3 outputs for FC4 and OP layers to match the dimension). Let $\zvec_i$ be one of those outputs in \eqref{eqn:notation.output.layer}. For those three classes that used for modeling for the frog data, we split them into the training set ($\Sset$) and the testing set ($\Tset$). The samples that we set aside for OOD testing is denoted by $\Oset$. We prepare three datasets for the NCD problem.
	
	\begin{inparaitem}
		\item For the training set, we collect $\{\zvec_i\}$ for those $i\in \Sset$, which is the output of a fully connected layer.
		
		\item For the in-distribution (ID) testing set, we feed the neural network with image $\Xb_i$ for those $i\in\Tset$ and then we collect $\{\zvec_i, u_i\}$ for those $i\in \Tset$. Here $u_i$ is the OOD sample indicator. For those $i\in \Tset$, we have $u_i=0$.
		
		\item For the OOD testing set, we feed the neural network with image $\Xb_i$ for those $i\in\Oset$ and then we collect $\{\zvec_i, u_i\}$ for those $i\in \Oset$. For those $i\in \Oset$, we have $u_i=1$.
		
	\end{inparaitem}
	
Then we assume $\zvec_i$ given class $j$ follows a multivariate normal distribution. That is, the distribution of $(\zvec|y=j)$ is modeled as  $\NOR(\zvec|\muvec_j,\Sigmavec_j)$, where the mean $\muvec_j$ is estimated by
\begin{align}\label{eqn:muhat.j}
\muvechat_j=&\frac{1}{n_j}\sum_{i \in \Cset_j\cap\Sset}\zvec_i.
\end{align}
Here $n_j$ is the number of samples in $\Cset_j\cap\Sset$ and $n$ is the number of samples in $\Sset$.

For the OOD detection, we consider both linear discriminate analysis (LDA) and quadratic discriminant analysis (QDA). For LDA, the variance-covariance matrix $\Sigma_j$ are estimated by
\begin{align}\label{eqn:Sigmahat.j}
\Sigmahat_j=\Sigmahat=&\frac{1}{n-1}\sum_j\sum_{i \in \Cset_j\cap\Sset}(\zvec_i-\muvechat_j)(\zvec_i-\muvechat_j)\tran,
\end{align}
which is common across all class labels. For QDA, each class has different $\Sigmahat_j$, which is
	\begin{align*}
		\Sigmahat_j=&\frac{1}{n_j-1}\sum_{i \in \Cset_j\cap\Sset}(\zvec_i-\muvechat_j)(\zvec_i-\muvechat_j)\tran.
	\end{align*}
Note that the estimate for the variance-covariance matrix, $\Sigmahat_j$, may not be positive definite. In such cases, we use the estimator from \citeN{LEDOIT2004365}, which works for non-positive definite cases by modifying the sample covariance matrix as,
\begin{align}\label{eqn:muhat.j}
	\Sigmahat_j^{\text{LW}}=\rho_1\Ivec_{n_j}+\rho_2\Sigmahat_j.
\end{align}
Here $\Ivec_{n_j}$ in \eqref{eqn:muhat.j} is the $n_j\times n_j$ identical matrix, and $\rho_1$ are $\rho_2$ are scalars. Note that $\rho_2\Sigmahat_j$ is a nonnegative definite matrix and adding $\rho_1\Ivec_{n_j}$ to its diagonal terms will make  $\Sigmahat_j^{\text{LW}}$ positive definite.

Specifically, $\rho_1$ are $\rho_2$ are determined by the following optimization problem,
\begin{align}\label{eqn:rho1.rho2.optim}
(\rho_1,\rho_2)\tran=\argmin_{\rho_1,\rho_2}\E
\Vert\Sigmahat_j^{\text{LW}}-\Sigmavec_j\Vert^2.
\end{align}
Here $\Vert\cdot\Vert^2$ is the squared Frobenius norm for matrices, and the expectation $\E(\cdot)$ is taken with respect to the distribution $\NOR(\zvec|\muvec_j,\Sigmavec_j)$. The idea behind \eqref{eqn:rho1.rho2.optim} is to find a matrix that is guaranteed to be positive definite and it is as close to $\Sigmavec_j$ as possible.

Intuitively, those $\zvec_i$'s from the OOD samples should be far away from the center of distribution of $\zvec$, while those $\zvec_i$'s from the ID samples should be close to  the center of distribution of $\zvec$, Thus, we use the Mahalanobis distance-based confidence score of $\zvec_i$ to measure the distance of $\zvec_i$ to its nearest class (e.g., \shortciteNP{Denouden2018ImprovingRA}). That is, for LDA, we compute
	\begin{align*}
		w_i=\min_j\; (\zvec_i-\muvechat_j)\tran\Sigmahat^{-1}(\zvec_i-\muvechat_j), i\in\Tset\cup\Oset,
	\end{align*}
and for QDA, we compute
	\begin{align*}
		w_i=\min_j\; (\zvec_i-\muvechat_j)\tran\Sigmahat_j^{-1}(\zvec_i-\muvechat_j), i\in\Tset\cup\Oset.
	\end{align*}

For a sample with $w_i$ that is beyond a threshold, it will classified as a new class. So the data for the NCD problem is denoted as $\{w_i, u_i\}, i\in\Tset\cup\Oset$.  We train a logistic regression model as
\begin{align}
u_i\sim \text{Bernoulli}(\xi_i), \quad \xi_i=\pr(u_i=1)=\frac{\exp(\alpha_0+\alpha_1 w_i)}{1+\exp(\alpha_0+\alpha_1 w_i)},
\end{align}
where $\xi_i$ is the probability. With the logistic regression and the area under the receiver operating characteristic curve (AUC), a threshold can be determined. Note that to learn the logistic model, training samples from the OOD set are needed.

%%%%%%%%%%%%%%%%%%%%%%%%%%%%%%%%%%%%%%%%%%%%%%%%%%%%%%%%%%%%%%%
\section{Data Analysis Results}\label{sec:results}
%%%%%%%%%%%%%%%%%%%%%%%%%%%%%%%%%%%%%%%%%%%%%%%%%%%%%%%%%%%%%%
In this section, we presents numeric results for the frog data by applying the methods described in Section~\ref{sec:method.development}. Because the frog dataset is relatively small, we also use a larger dataset, namely the MNIST data (\citeNP{Deng2012}), for illustration of the NCD methods and providing more insights.

%%%%%%%%%%%%%%%%%%%%%%%%%%%%%%%%%%%%%%%%%%%%%%%%%%%%%%%%%%%%%%
\subsection{Model Testing Results on Frog Data}\label{sec:frog.model.results}
%%%%%%%%%%%%%%%%%%%%%%%%%%%%%%%%%%%%%%%%%%%%%%%%%%%%%%%%%%%%%%

We first fit the CNN model as in Figure~\ref{fig:FrogCNN} to all the frog image data as summarized in Table~\ref{tab:frog.counts}. We use balanced cross validation as described in Section~\ref{sec:CNN.classification} with $K=5$, which leads to 80\% for training and 20\% for testing. Table~\ref{tab:CV} shows the accuracy of each fold and the overall accuracy for the testing set is around 73.1\%, which is reasonably good, given that the size of training set is only around 150. To gain insights on mis-classifications, Table~\ref{fig:CVconf} shows the overall confusion matrix. We can see that the model mainly mis-classifies clades 4 and 5 to other clades, which may be caused by limited sample size.
	
\begin{figure}[ht]%
		\begin{center}
			\begin{tabular}{cc}
				\includegraphics[width=.38\textwidth]{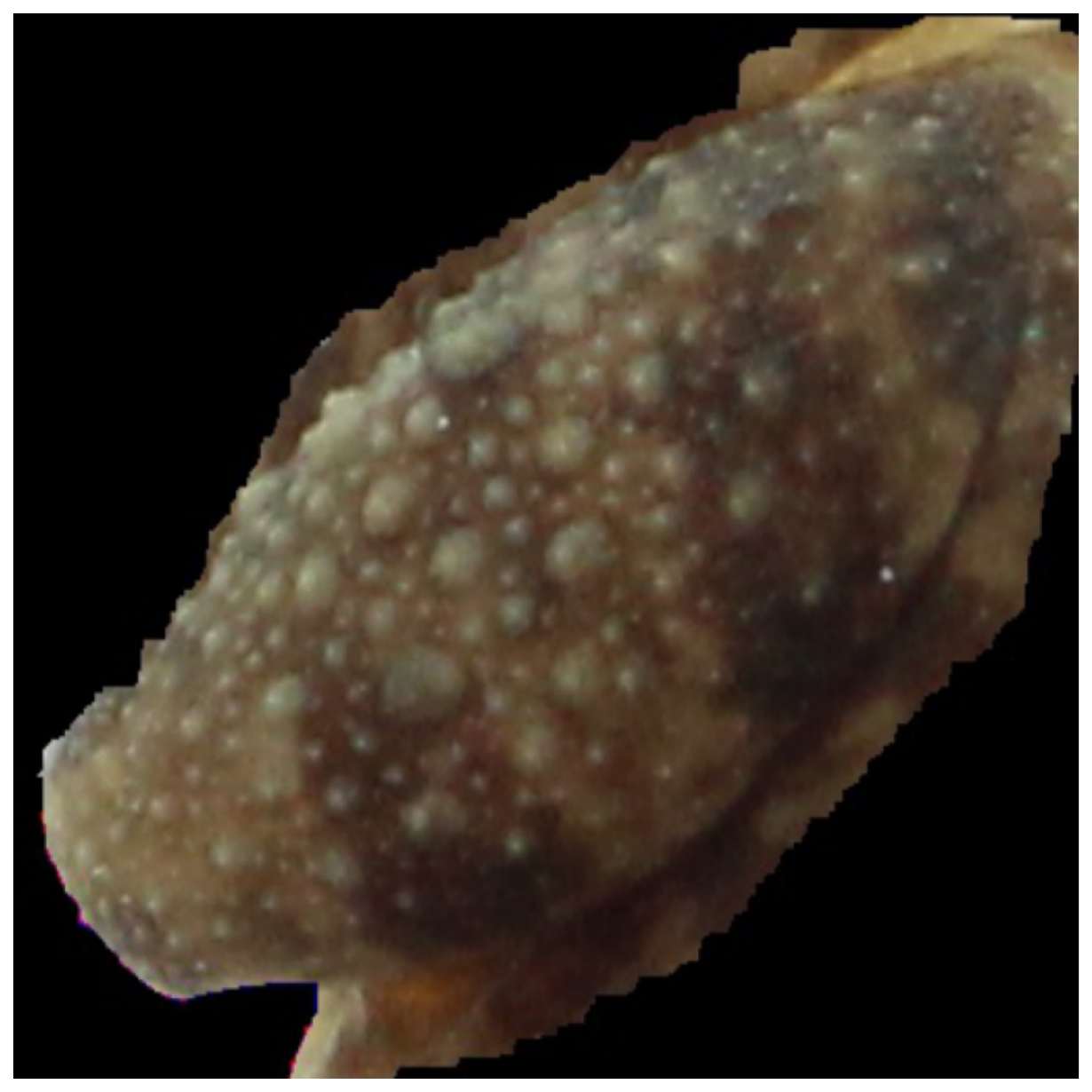}&
				\includegraphics[width=.39\textwidth]{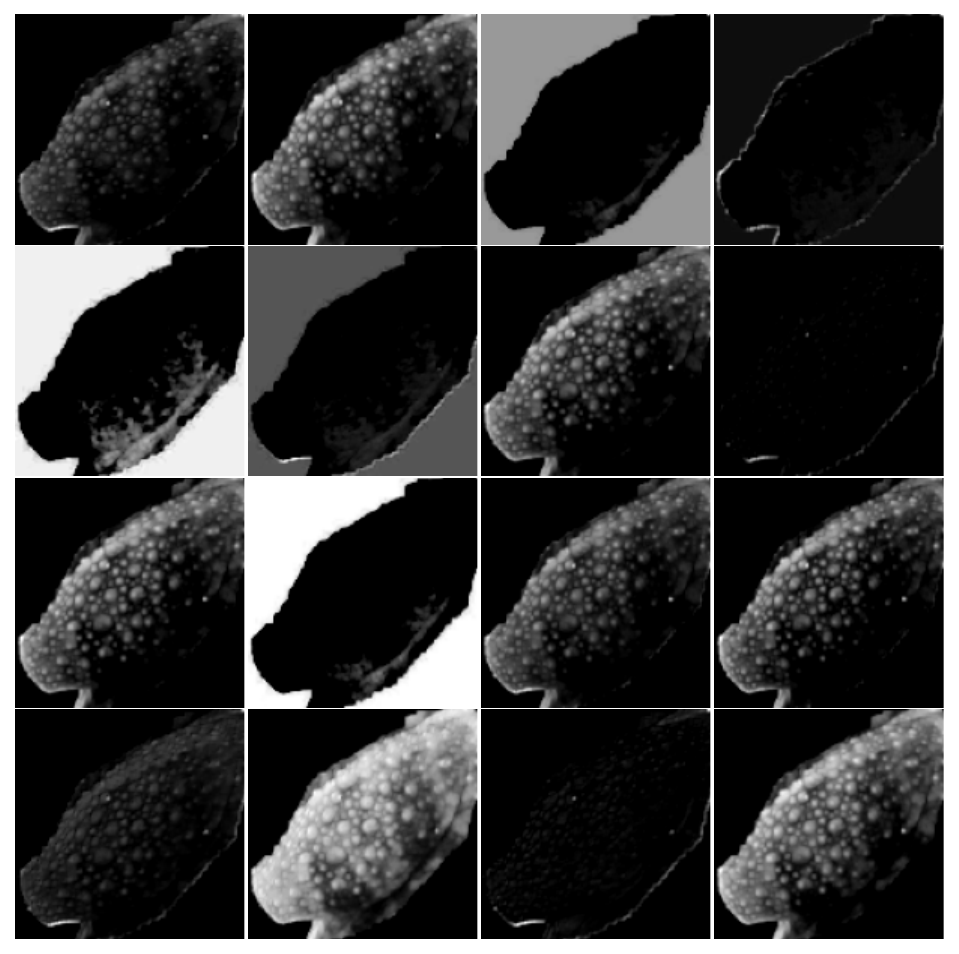} \\
                (a) Input layer & (b) Conv2: 16-unit layer\\
				\multicolumn{2}{c}{\includegraphics[width=0.79\textwidth]{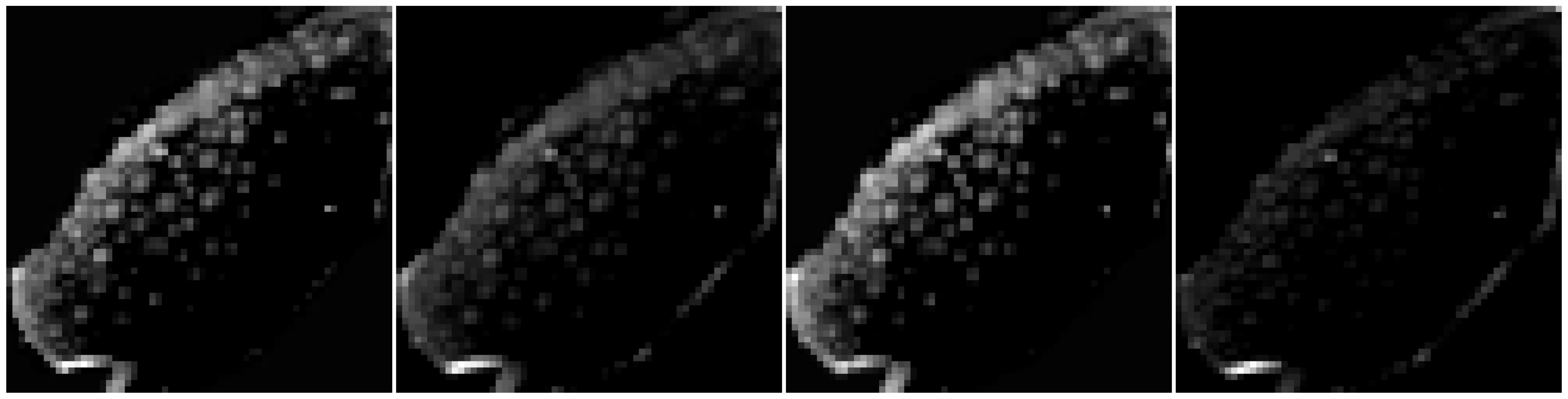}}\\
\multicolumn{2}{c}{ (c) Conv3: 4-unit layer}
			\end{tabular}
		
		\caption{Visualization of outputs of convolution layers in the CNN model for the frog images.}\label{fig:frog.cnn.layers}
\end{center}	
\end{figure}

We also plot some output of intermediate layers of the CNN model. Figure~\ref{fig:frog.cnn.layers} plots an input image from one frog and the output of the last two convolution layers (i.e., Conv2 and Conv3) in the CNN model from the input frog image. We can see that different filters pick up different features from the image. Especially from Figure~\ref{fig:frog.cnn.layers}(c), we can see that the tuberculation of frog leg are activated in the CNN model, which provides insight that the tuberculation can contribute to classification of frog clades.

	\begin{table}
		\begin{center}
		\caption{Accuracy for five cross validation folds for the frog data.}\label{tab:CV}
		\begin{tabular}{ccccccc}\hline\hline
			Cross Validation Fold&1		&2		&3		&4		&5 &Overall\\\hline
			Accuracy		&0.775	&0.675	&0.789	&0.711	&0.703&0.731\\\hline\hline
		\end{tabular}
        \end{center}
	\end{table}

	\begin{table}
		\begin{center}
		\caption{Confusion matrix for the five-fold cross validation for the frog data.}\label{fig:CVconf}
		\begin{tabular}{c|c|cccc}\hline\hline
			& \multicolumn{5}{c}{Predicted} \\\hline
			\multirow{5}{*}{True}&  & Clade 4   & Clade 5   & Clade 8 &  Clade 12 \\\cline{2-6}
			&   Clade 4 &  27 & 0  &  6  &   11\\
			&   Clade 5 &  1  & 9  &  10 &   2 \\
			&   Clade 8 &  1  & 2  &  70 &   4 \\
			&  Clade 12 &  8  & 0  &  6  &   35\\
\hline\hline
		\end{tabular}
	    \end{center}
	\end{table}

%%%%%%%%%%%%%%%%%%%%%%%%%%%%%%%%%%%%%%%%%%%%%%%%%%%%%%%%%%%%%%%%%%%%%%%%%%%%
\subsection{OOD Testing Results on Frog Data with One OOD Clade}
\begin{figure}%
		\begin{center}
			\begin{tabular}{cc}
				\includegraphics[width=.45\textwidth]{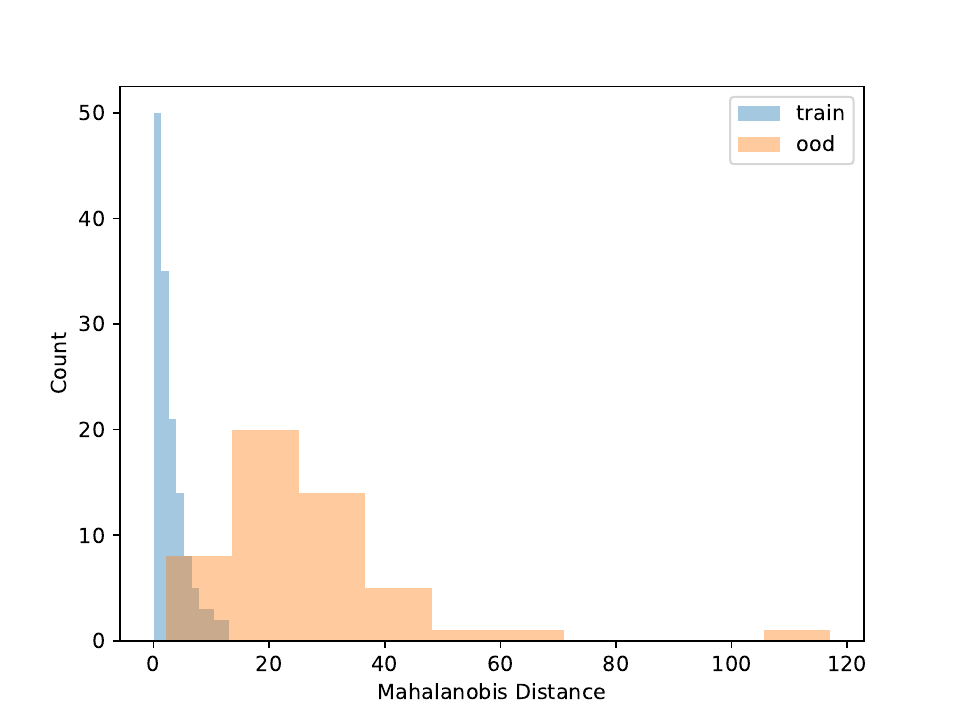}&
				\includegraphics[width=.45\textwidth]{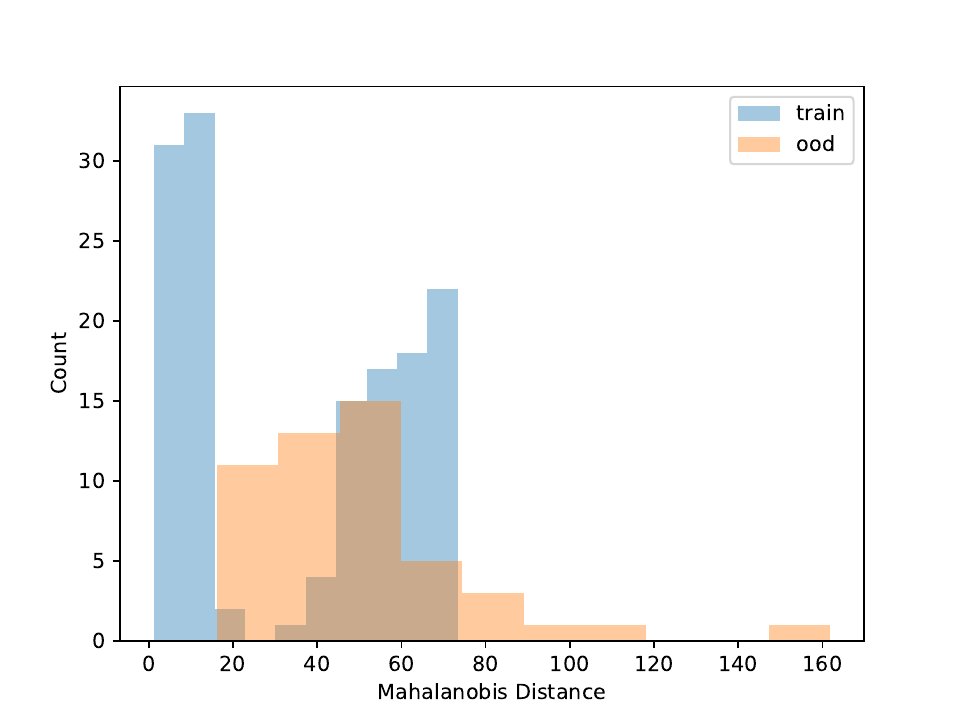} \\
                (a) FC4: 3-unit layer & (b) FC3: 64-unit layer\\		\multicolumn{2}{c}{\includegraphics[width=.45\textwidth]{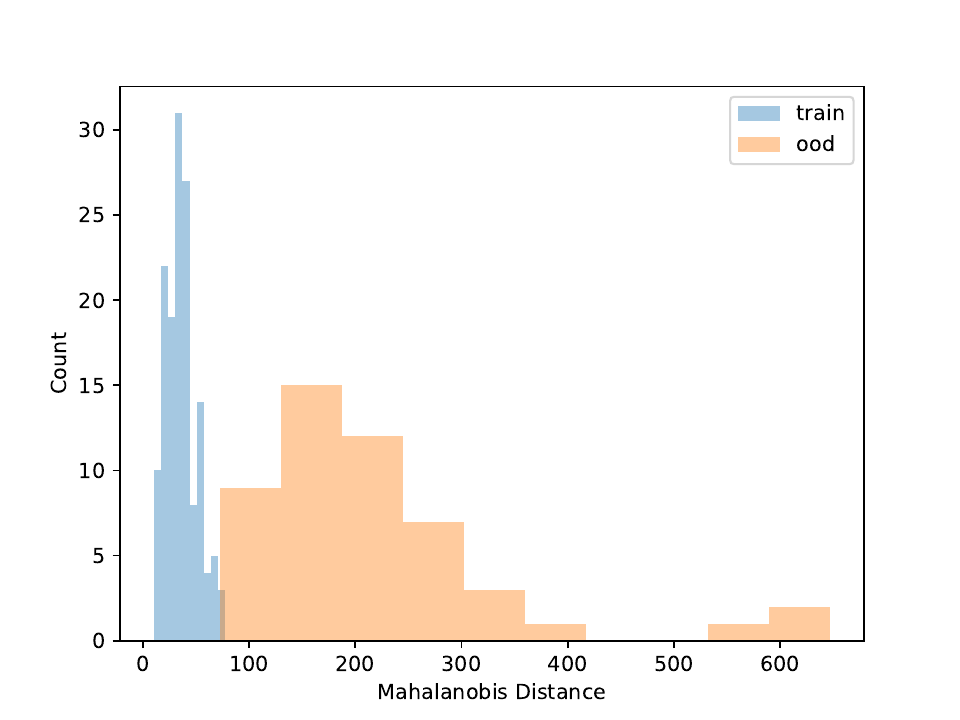}}\\
\multicolumn{2}{c}{ (c) FC1: 128-unit layer}
			\end{tabular}
		\caption{Distribution of Mahalanobis distance based on the frogs data using outputs from the fully connected layers. Clade 12 is treated as OOD samples.}\label{fig:froghist}
\end{center}
	\end{figure}

%%%%%%%%%%%%%%%%%%%%%%%%%%%%%%%%%%%%%%%%%%%%%%%%%%%%%%%%%%%%%%%%%%%%%%%%%%%%
Using the CNN model built in Section~\ref{sec:frog.model.results}, in this section, we conduct some OOD detections for the frog data with one OOD clade. The data splitting strategy is as described in Section~\ref{sec:NCD}. That is, we set one clade as the OOD sample and the rest to build the CNN model. Table~\ref{tab:frogsmetric} shows the fitting metrics based on the LDA on the frog data, with each species left out as OOD in each column. The accuracy of detecting OOD samples is high but also depends on the OOD sample size. Generally, more OOD samples will have satisfactory OOD accuracy. Figure~\ref{fig:froghist} shows the visualization of the NCD for clade 5. We can see that the distribution of known species and new species is different. But for the layers with 3 and 64 units, there are no major gaps between the two distributions, which shows it is challenging to do NCD. One reason could be due to the small sample sizes of the OOD sets. Because of the limited sample size, we do not fit the QDA model because the covariance matrix can not be robustly estimated for each class.

	\begin{table}
\begin{center}	
		\caption{The data for the CNN model building and the AUC when the species is treated as the OOD species, together with the confusion matrix. The total number of images is 193.}\label{tab:frogsmetric}
		\begin{tabular}{ccccc}\hline\hline
			OOD Clade & Clade 4 & Clade 5 & Clade 8 & Clade 12 \\
			\hline
			Number of Legs		&		44	&	22		&	77		&	50	\\\hline
Model Building Clades & 5, 8, 12 & 4, 8, 12 & 4, 5, 12 & 4, 5, 8 \\\hline
			AUC &0.963 &0.791 & 0.969 & 0.999\\\hline\\[-1ex]
			Confusion matrix &
			$\left[ \begin{array}{cc} 149 & 0  \\ 8 & 36 \end{array}\right]$
			&
			$\left[ \begin{array}{cc} 171 & 0  \\ 8 & 14 \end{array}\right]$
			&
			$\left[ \begin{array}{cc} 114 & 2  \\ 12 & 65 \end{array}\right]$
			&
			$\left[ \begin{array}{cc}143 & 0  \\ 1 & 49 \end{array}\right]$\\[3ex]
			\hline\hline
		\end{tabular}
\end{center}
	\end{table}

%%%%%%%%%%%%%%%%%%%%%%%%%%%%%%%%%%%%%%%%%%%%%%%%%%%%%%%%%%%%%%%%%%%
\subsection{OOD Testing Results on Frog Data with Multiple OOD Clades}
\begin{figure}%
\begin{center}
\begin{tabular}{cc}
\includegraphics[width=.45\textwidth]{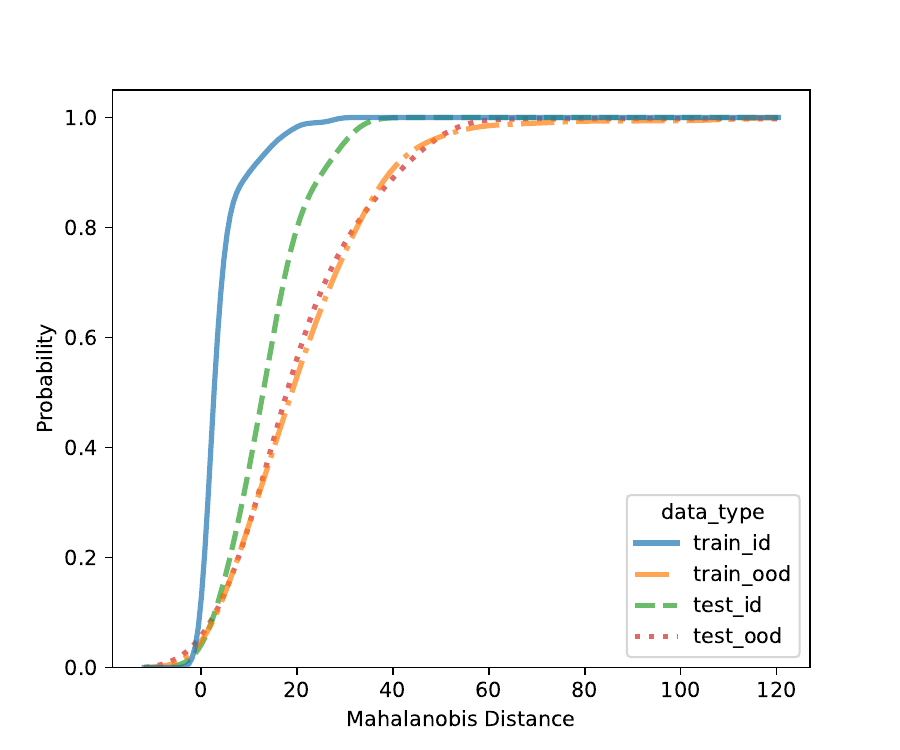}&
\includegraphics[width=.45\textwidth]{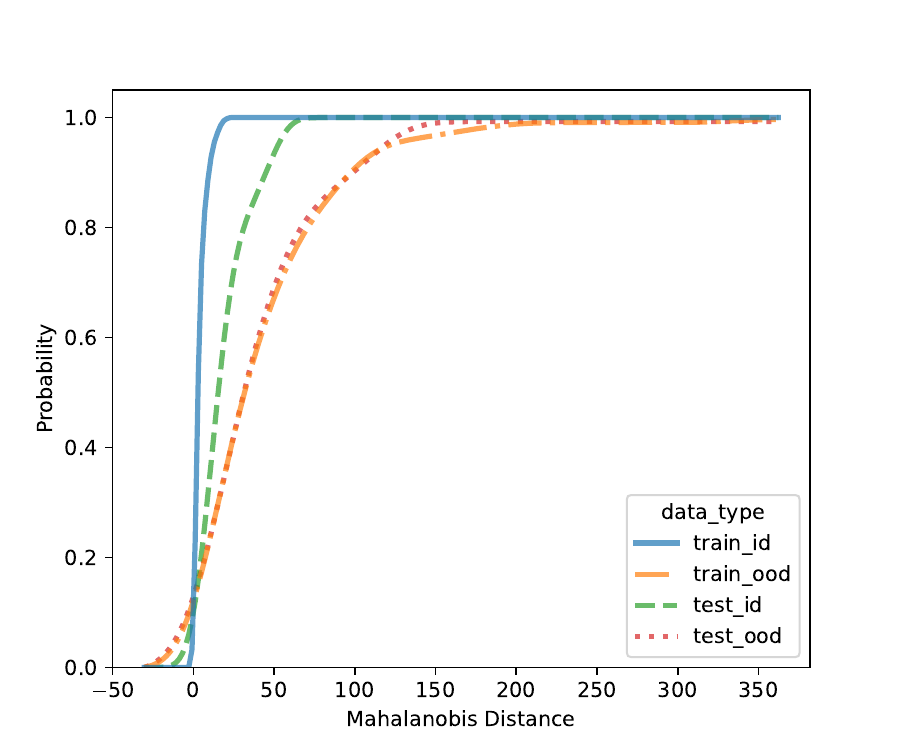}\\
(a) LDA based on FC4: 4-unit layer & (b) QDA based on FC4: 4-unit layer  \\
\includegraphics[width=.45\textwidth]{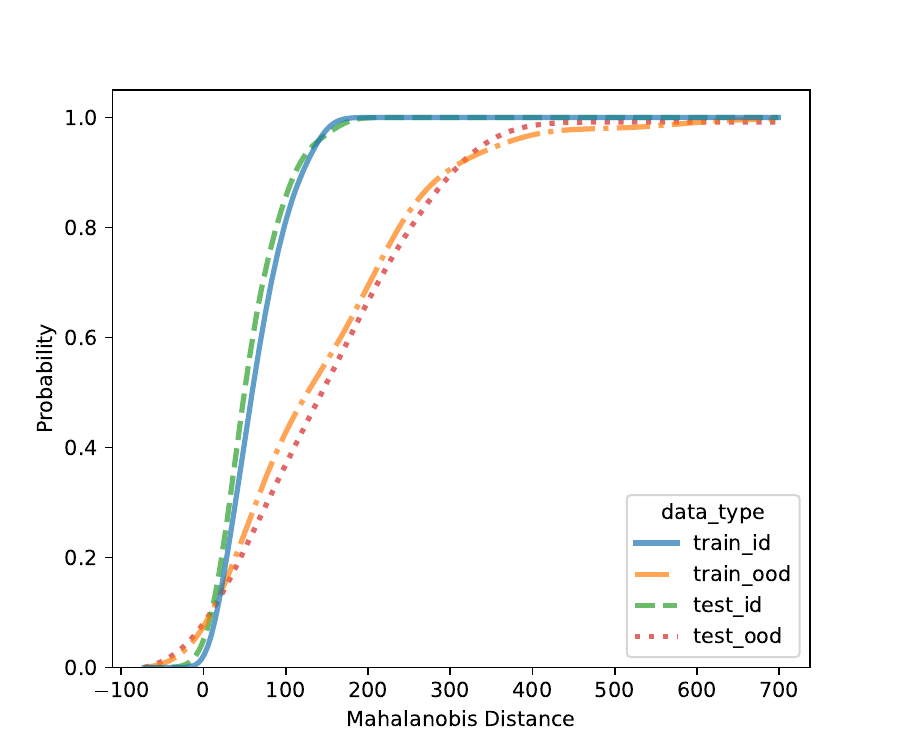} &
\includegraphics[width=.45\textwidth]{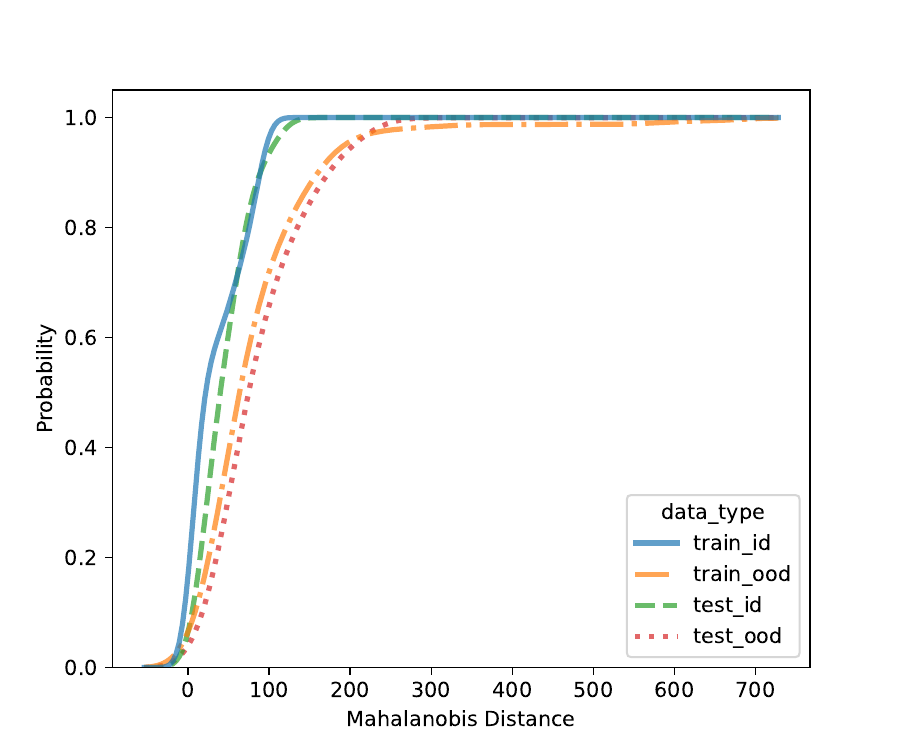} \\
(c) LDA based on FC3: 64-unit layer & (d) QDA based on FC3: 64-unit layer\\		
\includegraphics[width=.45\textwidth]{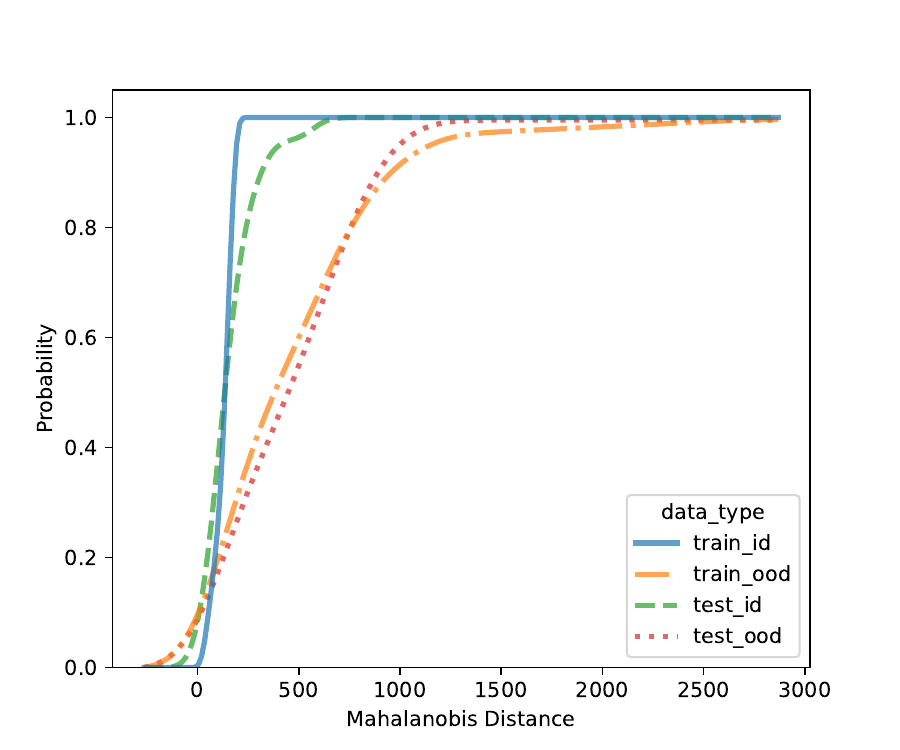}&
\includegraphics[width=.45\textwidth]{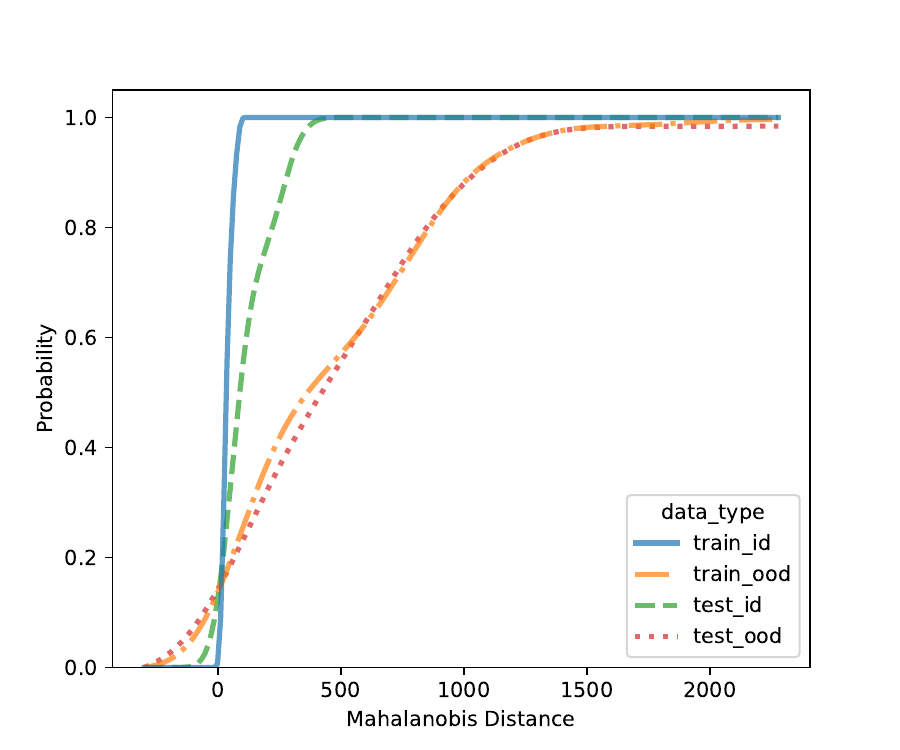}\\
(e) LDA based on FC1: 128-unit layer & (f) QDA based on QDA FC1: 128-unit layer
\end{tabular}
\caption{Cumulative distribution of Mahalanobis distance of different fully-connected layers for the LDA and QDA based on the frog data. Clades 4, 5, 8, and 12 are used as ID samples and clades 10, 11, 16, 18, 20, and 21 are treated as OOD samples.}\label{fig:frog.new.ood.lda.QDA}
\end{center}
\end{figure}

%%%%%%%%%%%%%%%%%%%%%%%%%%%%%%%%%%%%%%%%%%%%%%%%%%%%%%%%%%%%%%%%%%%
In this section, we present additional results on OOD detection because some new samples were gathered, especially for clades 10, 11, 16, 18, 20, and 21. Because of the availability of additional images on multiple new clades other than clades 4, 5, 8, and 12, we can form an OOD class from multiple OOD clades from those new clades. In particular, the leg image counts for those new clades are 24, 9, 40, 61, 28, and 27, respectively. In total, we have 189 leg images for the OOD class. This section uses images from clades 4, 5, 8, and 12 to train the CNN model, as shown in Figure~\ref{fig:FrogCNN}. Figure~\ref{fig:frog.new.ood.lda.QDA} shows the cumulative distribution of Mahalanobis distance of different fully-connected layers for the LDA and QDA methods. We see that the QDA procedure has slightly better classification power. Table~\ref{tab:frogsmetric.new.ood} shows the metrics for LDA and QDA for OOD detection based on the frog data. We can see that the QDA method has better overall performance when considering AUC, true positive rate (TPR), and true negative rate (TNR). Overall, the OOD detection ability is good based on the multiple OOD clade data, though there is still room for improvement.

\begin{table}
\begin{center}	
\caption{Metrics for LDA and QDA for OOD detection based on the frog data. Clades 10, 11, 16, 18, 20, and 21 are treated as OOD samples.}\label{tab:frogsmetric.new.ood}
\begin{tabular}{cccccccccc}\hline\hline
\multicolumn{5}{c}{LDA}	&  \multicolumn{5}{c}{QDA} \\\cline{2-4}\cline{7-9}
& Metric & Fitting & Prediction &&& Metric & Fitting & Prediction&\\\hline
&AUC &  0.975 & 0.747 &&& AUC & 0.912 & 0.750& \\
&TPR &  0.870 & 0.947 &&& TPR & 0.758 & 0.789& \\
&TNR &  0.961 & 0.208 &&& TNR & 0.942 & 0.416& \\\hline\hline
\end{tabular}
\end{center}
\end{table}

%%%%%%%%%%%%%%%%%%%%%%%%%%%%%%%%%%%%%%%%%%%%%%%%%%%%%%%%%%%%%%%%%%%%%%%%%%%%%%%%%%%%%%%%%%%%%%%%%%%%%%%%%%%%
\subsection{OOD Testing Results on the MNIST Data}
\begin{figure}[ht]
\begin{center}
		\includegraphics[width=.8\textwidth]{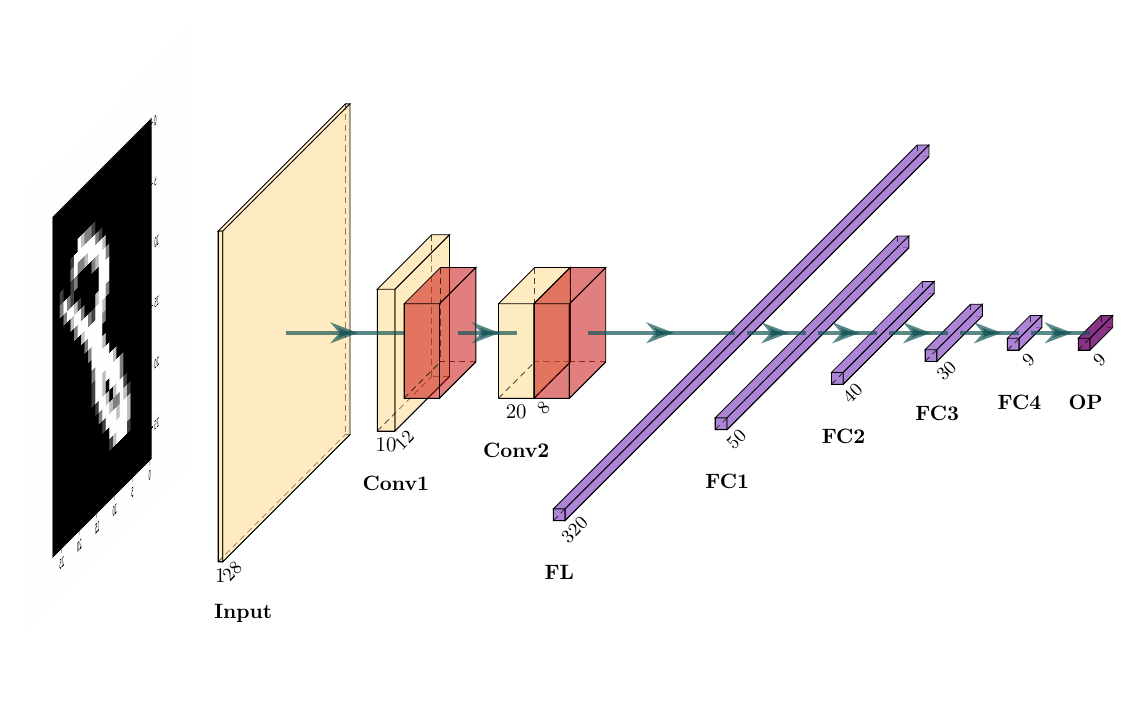}
		\caption{Diagram of CNN that is used for classification on the MNIST data. Here ``Conv1'' means convolution layer 1, ``MP" means max pooling, ``FL'' means flatten layer, ``FC1'' means fully connected layer 1, and ``OP'' means output layer.}\label{fig:MNISTCNN}
\end{center}
	\end{figure}

%%%%%%%%%%%%%%%%%%%%%%%%%%%%%%%%%%%%%%%%%%%%%%%%%%%%%%%%%%%%%%%%%%%%%%%%%%%%%%%%%%%%%%%%%%%%%%%%%%%%%%%%%%%%
To better study the performance of our NCD methods, we further test our method by using the much larger MNIST image dataset in \citeN{Deng2012}, which has tens of thousands of images. The MNIST data are images of handwritten numbers. Although the frog images and MNIST images look different, the general idea of using the features extracted from CNN to do the OOD detection can be applied to different types of image data. Because the input image of the MNIST data is different (much simpler to some extent) from the frog images, we need to slightly modify our CNN model. In particular, we use the CNN architecture as shown in Figure~\ref{fig:MNISTCNN}, with fewer layers than the CNN for the frog data.

\begin{figure}
\begin{center}
\begin{tabular}{ccc}
\includegraphics[width=0.2\textwidth, height=0.2\textwidth]{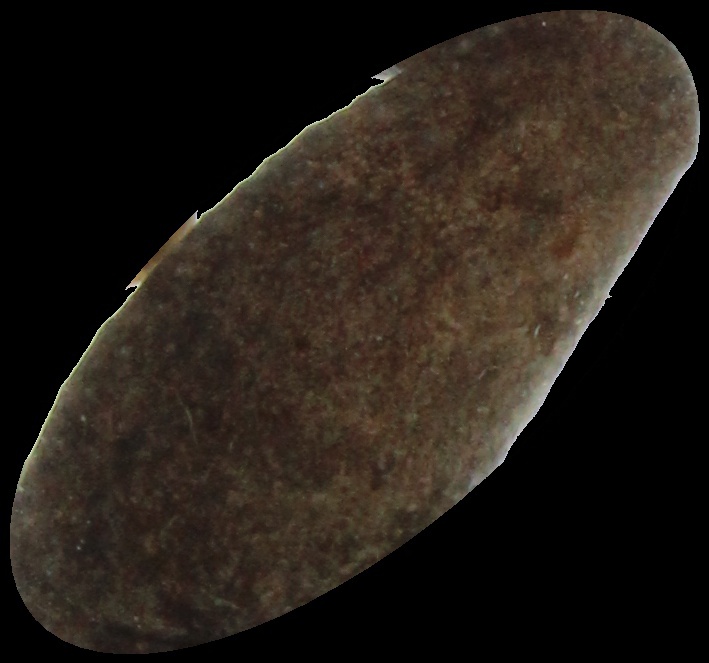} &
\includegraphics[width=0.2\textwidth, height=0.2\textwidth]{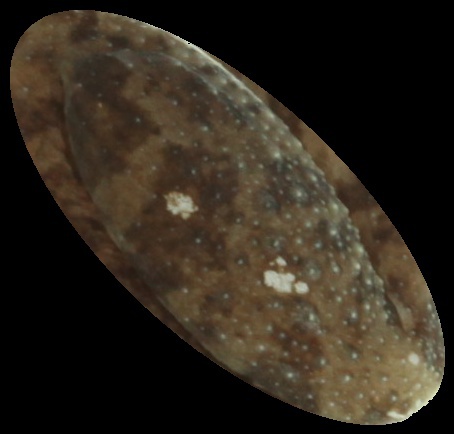} &
\includegraphics[width=0.2\textwidth, height=0.2\textwidth]{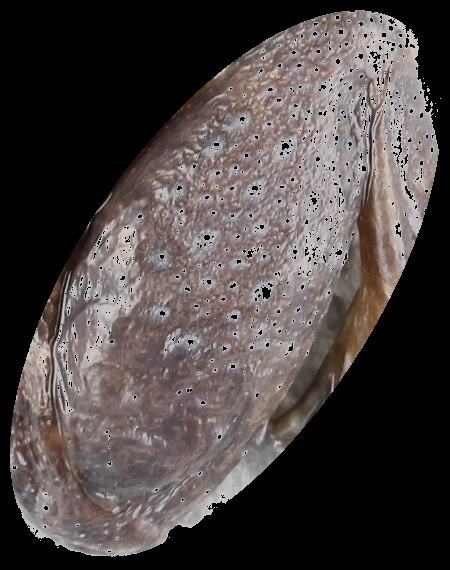} \\
(a) Clade 10   & (b) Clade 11  & (c) Clade 16   \\[1ex]
\includegraphics[width=0.2\textwidth, height=0.2\textwidth]{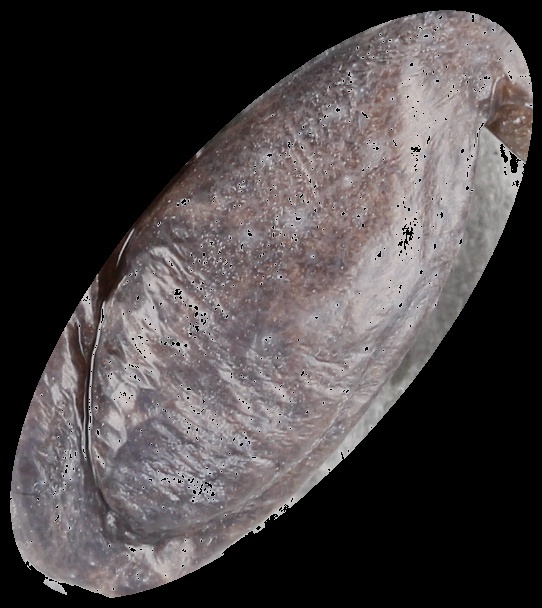} &
\includegraphics[width=0.2\textwidth, height=0.2\textwidth]{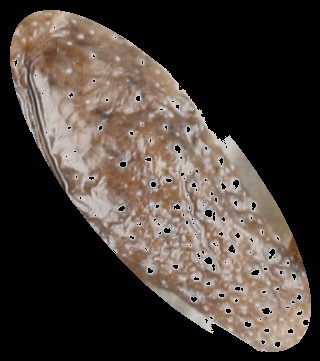} &
\includegraphics[width=0.2\textwidth, height=0.2\textwidth]{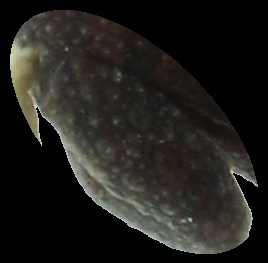} \\
(d) Clade 18  & (e) Clade 20   & (f) Clade 21  \\[1ex]
\end{tabular}			
\caption{Example leg images from the \emph{L. kuhlii} species that are treated as OOD samples.}\label{fig:new.ood.frog.leg.images}
\end{center}
\end{figure}

\begin{figure}%
		\begin{center}
			\begin{tabular}{cc}
				\includegraphics[width=.48\textwidth]{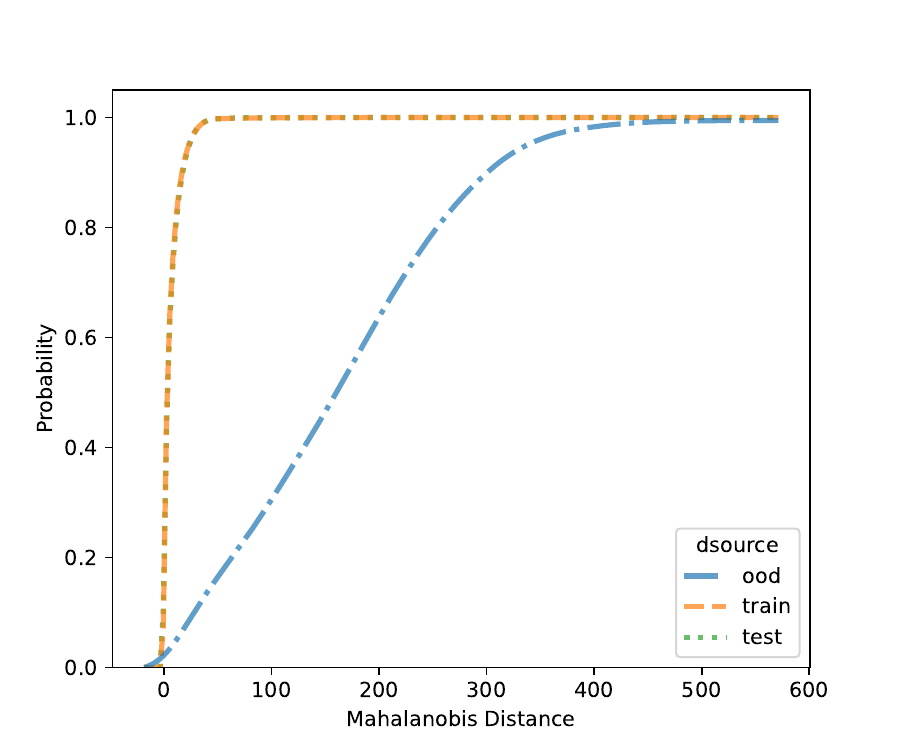}&
				\includegraphics[width=.48\textwidth]{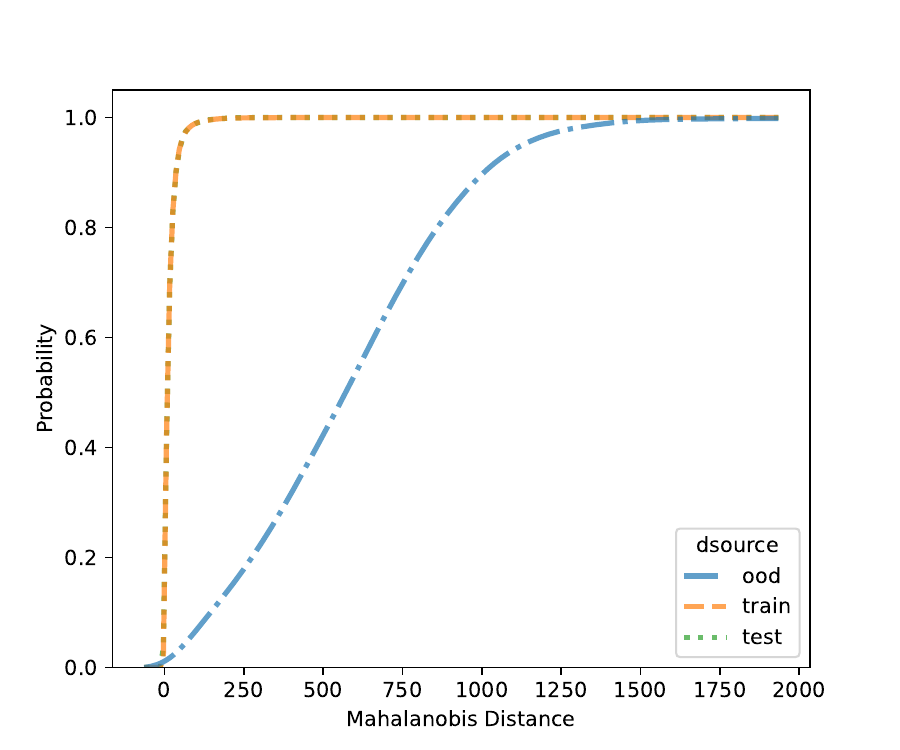} \\[-2ex]
				(a) FC4: 9-unit layer & (b) FC3: 30-unit layer \\
\includegraphics[width=.48\textwidth]{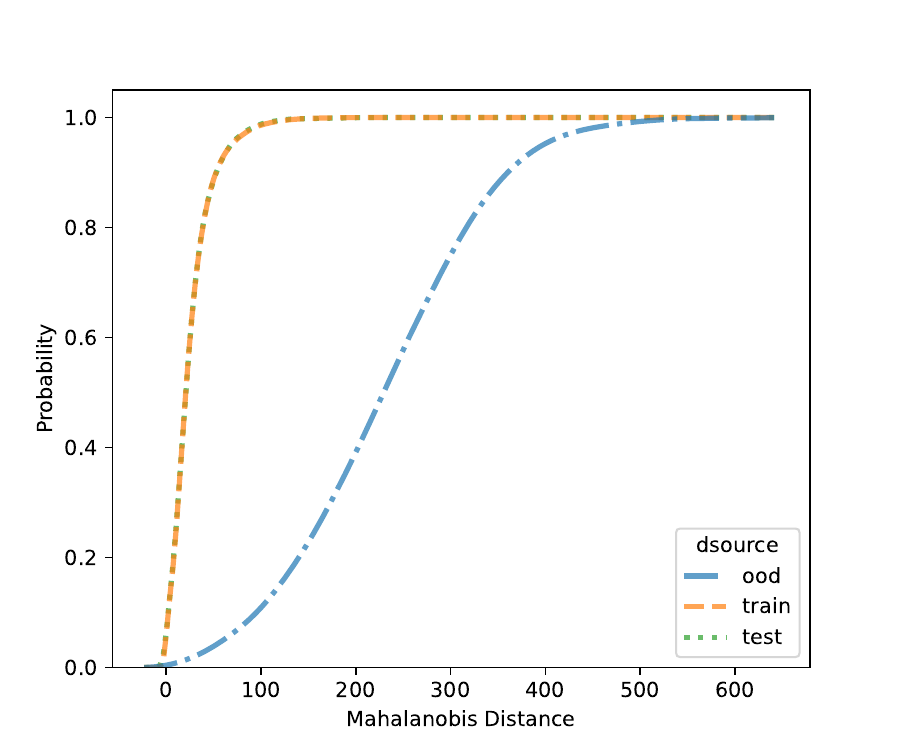}&
\includegraphics[width=.48\textwidth]{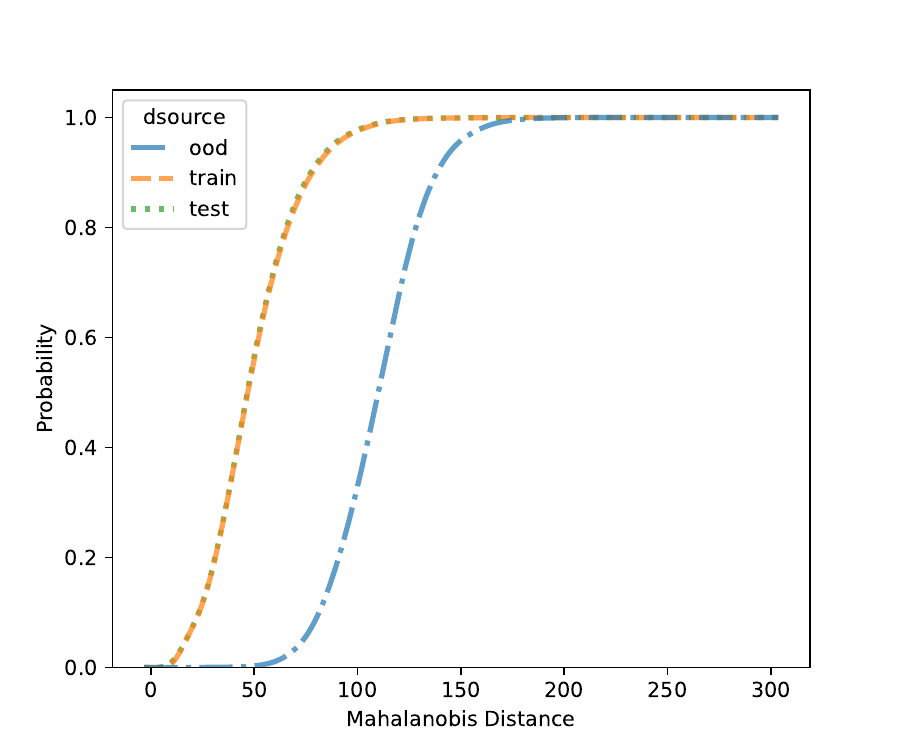}\\[-2ex]
				(c) FC2: 40-unit layer & (d) FC1: 50-unit layer \\
\multicolumn{2}{c}{\includegraphics[width=.48\textwidth]{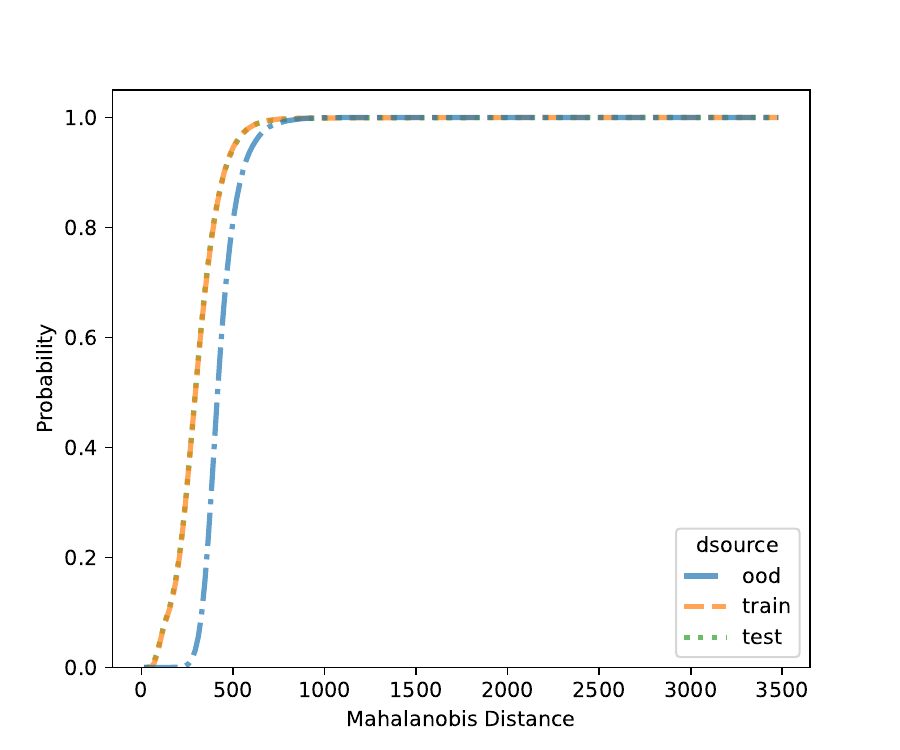}}\\[-2ex]
\multicolumn{2}{c}{(e) FL: 320-unit layer}
			\end{tabular}
		\caption{Cumulative distribution of Mahalanobis distance of different fully-connected layers for the LDA based on the MNIST data testing set, number ``9'' is treated as OOD samples.}\label{fig:MNISTLDA}
\end{center}
	\end{figure}

\begin{figure}%
		\begin{center}
			\begin{tabular}{cc}
				\includegraphics[width=.48\textwidth]{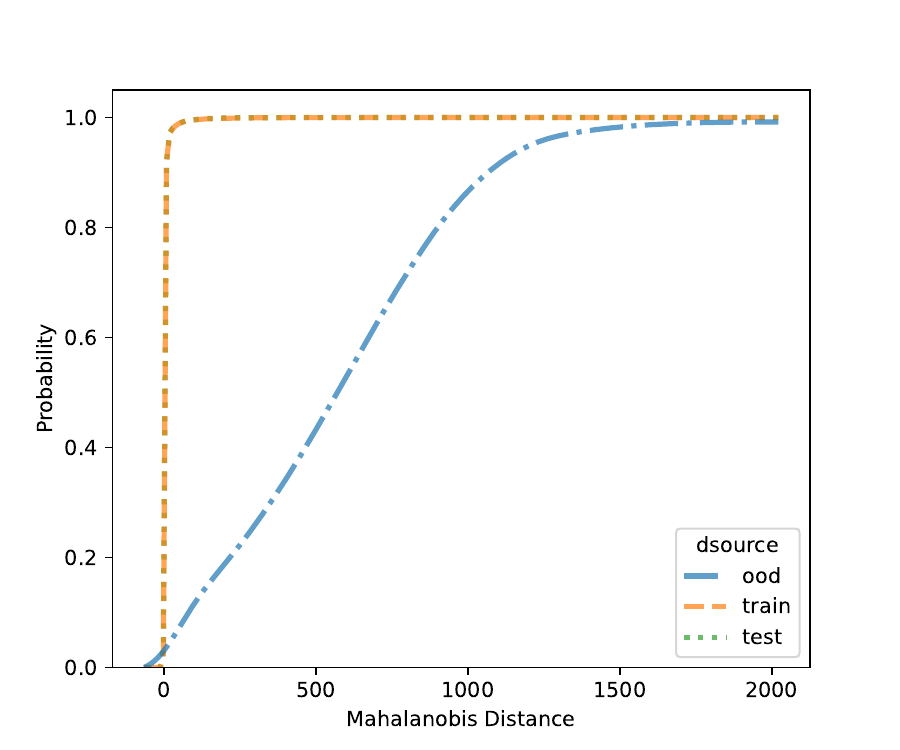}&
				\includegraphics[width=.48\textwidth]{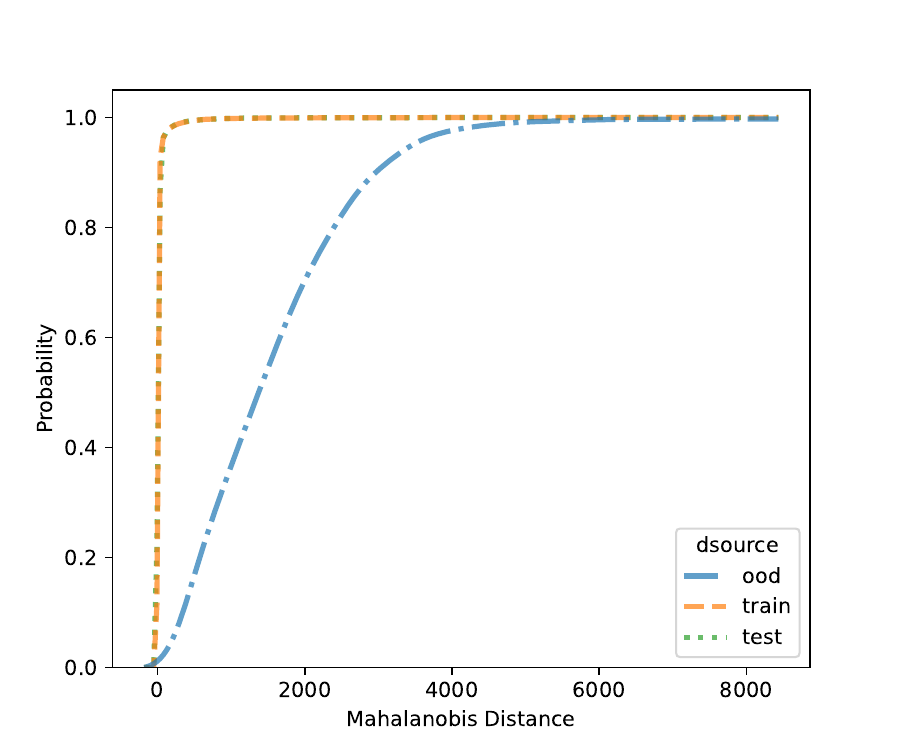}\\[-2ex]
(a) FC4: 9-unit layer & (b) FC3: 30-unit layer\\
\includegraphics[width=.48\textwidth]{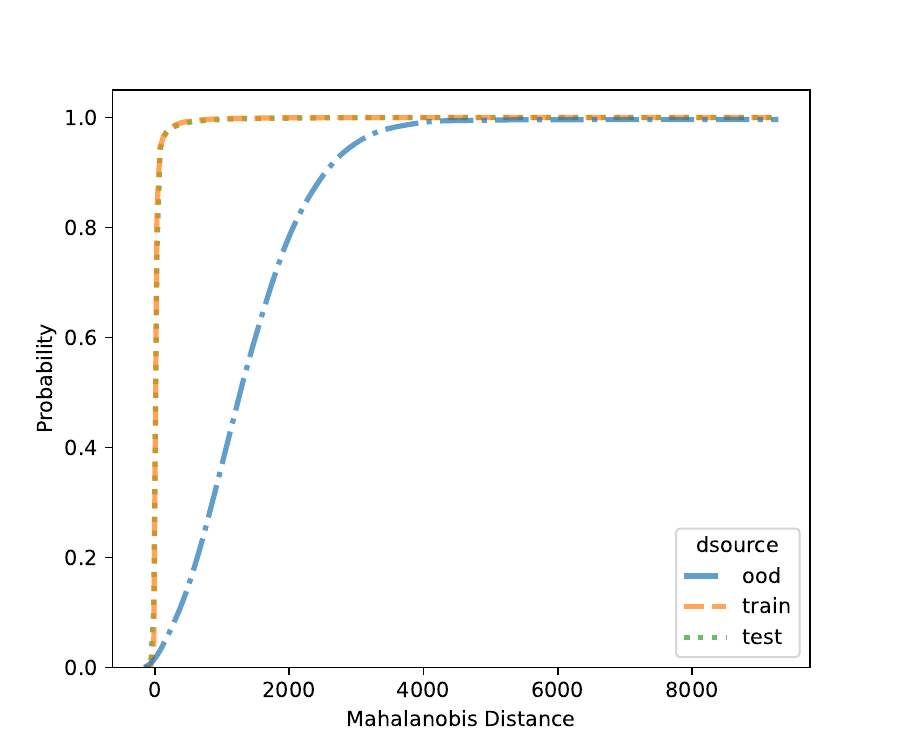}&
\includegraphics[width=.48\textwidth]{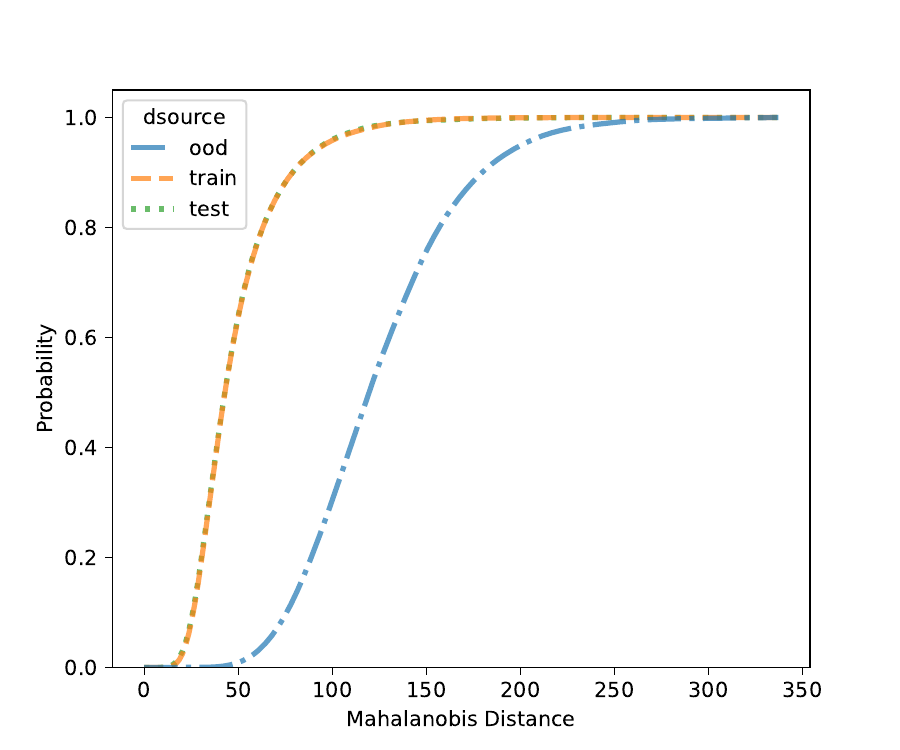}\\[-2ex]
				(c) FC2: 40-unit layer & (d) FC1: 50-unit layer\\
\multicolumn{2}{c}{\includegraphics[width=.48\textwidth]{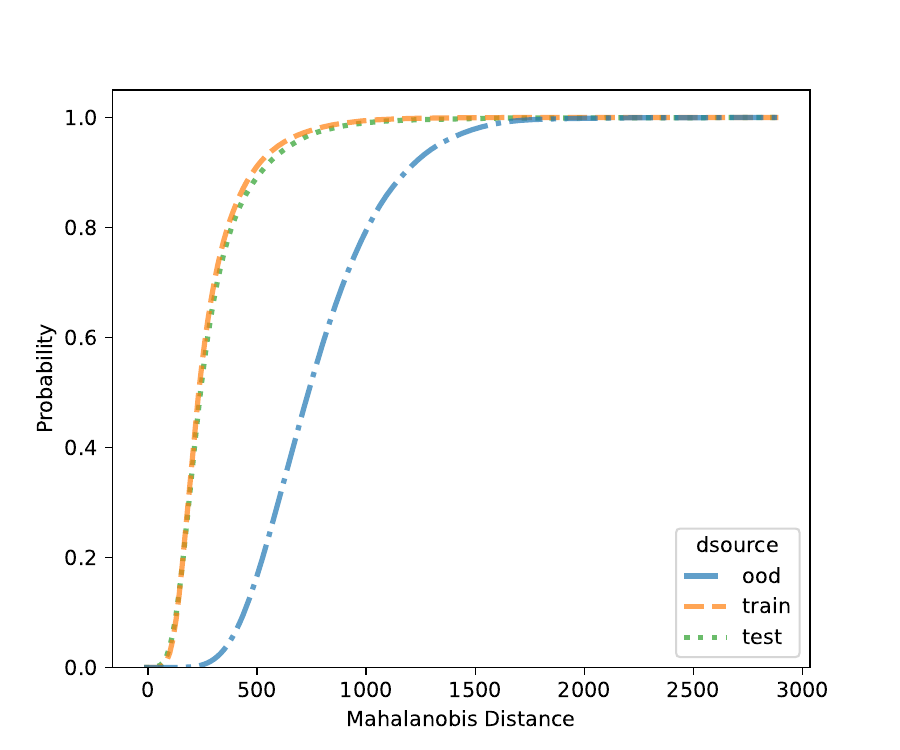}}\\[-2ex]
\multicolumn{2}{c}{(e) FL: 320-unit layer}
			\end{tabular}
		\caption{Cumulative distribution of Mahalanobis distance of different fully-connected layers for the QDA based on the MNIST data testing set, number ``9'' is treated as OOD samples.}\label{fig:MNISTQDA}
\end{center}

	\end{figure}

For NCD of the MNIST, we now can consider both classification methods, LDA and QDA. Figures~\ref{fig:MNISTLDA}~and~\ref{fig:MNISTQDA} shows the distribution of the Mahalanobis distance using the MNIST data based on LDA and QDA where the number ``9'' is left out as OOD. The distribution of the layer's output of OOD is significantly different from the training and testing data. We see that the QDA procedure has better classification power on layers with 320 and 50 outputs while LDA is better on the final fully connected layer with 8 outputs. One simple idea is when building a logistic model, use the QDA Mahalanobis distance from the layers with 320 and 50 outputs, and the LDA distance from the layer with 8 outputs. Table~\ref{tab:MNISTLDA.MNISTQDA} shows the accuracy results for the LDA and QDA in terms of AUC, TPR, and TNR, based on the MNIST data. We can see that both LDA and QDA have good accuracy for detecting new classes, and QDA has better prediction accuracy than the LDA especially as measured by TPR. In addition, QDA's performance is more stable when the OOD classes are changing.

	\begin{table}
		\begin{center}
			\caption{Metrics for LDA and QDA for different OOD samples, based on the MNIST data. }\label{tab:MNISTLDA.MNISTQDA}
			\begin{tabular}{c|c|cccccccccc}
				\hline\hline
\multicolumn{2}{c|}{OOD Class} &  0 &  1 &  2 &  3 &  4 &  5 &  6 &  7 &  8 &  9 \\\hline
\multicolumn{2}{c|}{CNN Accuracy} & 0.980 &	0.971 &	0.980&	0.733&	0.979&	0.867&	0.976&	0.928&	0.974&	0.984\\\hline
\multirow{3}{*}{LDA}	&AUC &  0.973 &  0.989 &  0.988 &  0.986 &  0.983 &  0.993 &  0.997 &  0.990 &  0.993 &  0.996 \\
				&TPR &  0.581 &  0.825 &  0.774 &  0.843 &  0.786 &  0.877 &  0.943 &  0.819 &  0.875 &  0.893 \\
				&TNR &  0.979 &  0.987 &  0.988 &  0.990 &  0.994 &  0.995 &  0.998 &  0.993 &  0.994 &  0.996 \\\hline
\multirow{3}{*}{QDA}
				&AUC &  0.991 &  0.993 &  0.991 &  0.988 &  0.996 &  0.992 &  0.999 &  0.994 &  0.998 &  0.993 \\
				&TPR &  0.855 &  0.851 &  0.843 &  0.832 &  0.921 &  0.877 &  0.953 &  0.888 &  0.964 &  0.880 \\
				&TNR &  0.984 &  0.992 &  0.985 &  0.985 &  0.993 &  0.988 &  0.998 &  0.994 &  0.994 &  0.994 \\\hline\hline
			\end{tabular}
		\end{center}
\end{table}

%%%%%%%%%%%%%%%%%%%%%%%%%%%%%%%%%%%%%%%%%%%%%%%%%%%%%%%%%%%%%%%%%%%%%%%%%%%%%%%%%%%%%%%%%%%%%%%%%%%%%%%%%%%%%%%%%
	\section{Concluding Remarks}\label{sec:conluding.remarks}
	%%%%%%%%%%%%%%%%%%%%%%%%%%%%%%%%%%%%%%%%%%%%%%%%%%%%%%%%%%%%%%%%%%%%%%%%%%%%%%%%%%%%%%%%%%%%%%%%%%%%%%%%%%%%%%%%%

In order for biodiversity conservation to be comprehensively successful we must be able to account for the species we are trying to protect. It is estimated that 86\% of existing species on Earth and 91\% of species in the ocean still await description (IUCN, www.iucn.org). Although progress towards species discovery, description, and conservation continues to be made, new tools are needed that will allow us address current and future challenges and impediments. Species complexes contribute to this taxonomic backlog because our current tools and knowledge are inadequate to completely resolve them. In an effort to address this, we applied deep learning methods to the morphologically similar, but genetically diverse group of frogs from Southeast Asia referred to the \emph{Limnonectes kuhlii} complex.

Our first objective was to use the species-specific pattern of skin tuberculation in these frogs as the detection criterion. We chose this morphological feature intentionally to see if a traditionally qualitative character could be treated quantitatively, thereby eliminating the subjectivity of the human observer. Our second objective was to present the model with an ``unknown'' image and see if it could correctly detect it as a new class (viz., a potentially new and undescribed species). We endeavored to achieve these objectives using a small sample set (193 images representing four classes).

Our results suggest that a CNN model for image classification can achieve good accuracy in identifying a species based on observable, morphological characters. This is especially significant considering our use of a single trait and a limited sample size.  Our model also succeeded in capturing a qualitative trait (tuberculation) and evaluating it in a consistent and quantitative manner. The ability to reliably and consistently quantify such a characteristic greatly reduces observer bias and provides a method for collecting standardized data that can be subjected to statistical analyses when evaluating the degree of similarity between two sample groups as is often done in traditional studies of comparative morphology. In an effort to increase the robustness of the CNN model in this application, future iterations could include additional morphological characteristics (e.g., head shape as suggested by \citeNP{Schoen2020}, and \shortciteNP{Stuartetal2020}) and geospatial data. Ideally, future studies will expand the taxonomic breadth of the model to include all known species in the \emph{L. kuhlii} complex and increase the sample size of images upon which the model is based. We also envision our model's framework being applicable to species complexes within any taxon.

With respect to our second objective, our NCD algorithms show effectiveness in recognizing an unknown species, especially considering the limitations of our sample size. Because of the uniqueness of our application, it takes a significant amount of time and budgets to collect frog image data at a large scale. The small sample size problem also prevents us from using famous CNN structures such as VGG (e.g., \citeNP{vgg16}). Some research, however, suggests that CNN can also work well if the CNN structure is carefully chosen (e.g., \shortciteNP{Dsouzaetal2020}). We expect the accuracy of our procedure to increase as the sample size increases, as demonstrated using the larger MNIST data. The ability to automate the detection of new species while simultaneously providing quantitative character-based evidence for the validity of the species would significantly facilitate and accelerate the processes of taxonomy.

Out research also suggests the need for establishing a public data repository for images. Considering that such repositories already exist (e.g, iNaturalist contains more than 70 million images and 138 million observations, www.inaturalist.org/observations), it is conceivable that these, or others like them could provide the data from which more accurate deep-learning models can be trained. For challenges such as those presented by species complexes, coordinated efforts among members of taxonomic working groups and museums could facilitate this by sharing open source data and images.

At present, anyone with a mobile electronic device (i.e., phone, tablet, etc.) can identify the species that commonly occur around them using mobile apps such as iNaturalist. This ability is facilitated by a platform that utilizes a computer vision model and an image set compiled by citizen scientists around the globe. However, corrections of species identifications still relies on a community of taxonomic experts who can visually vet the images and correct mistakes in identifications. App tools, such as Wildlife Spotter (https://wildspotter.org/),  are excellent and have resulted in new species detection and discovery.  We envision the refinement of those app tools with machine-learning methods, such as those developed in this paper, would have the ability to provide fine scale species resolution and the ability to correctly identify the members of a species complex. Moreover, we envision these tools to be capable of detecting and provide alerts when a potentially undescribed species is encountered.

Beyond our primary goals of species identification and detection, we also recognize that our results suggest that future research should explore the quantification of uncertainty in classification and NCD.  With uncertainty quantification, our NCD methods could also be used for estimating detection probability for species abundance estimates (e.g., \shortciteNP{Martin-Schwarzeetal2017}), or determining whether a prediction is reliable (e.g., \shortciteNP{Hongetal2022}) or robust (e.g., \shortciteNP{Lianetal2021Robustness}) for AI algorithms.

%

%%%%%%%%%%%%%%%%%%%%%%%%%%%%%%%%%%%%%%%%%%%%%%%%%%%%%%%%%%%%%%%%%%%%%%%%%%%%%%%%%%%%%%%%%
\section*{Supplementary Materials}

The following supplementary materials are available online.

\begin{description}
\item[Code and data:] Computing code and the frog leg image data are available at GitHub repository ``FrogLegImages''. The link to the repository is  \url{https://github.com/tgh1122334/FrogLegImages}.
	
\end{description}

%%%%%%%%%%%%%%%%%%%%%%%%%%%%%%%%%%%%%%%%%%%%%%%%%%%%%%%%%%%%%%%%%%%%%%%%%%%%%%%%%%%%%%%%%%%%%%%%%%%%%%%%%%%%%%%%%
\section*{Acknowledgments}
%%%%%%%%%%%%%%%%%%%%%%%%%%%%%%%%%%%%%%%%%%%%%%%%%%%%%%%%%%%%%%%%%%%%%%%%%%%%%%%%%%%%%%%%%%%%%%%%%%%%%%%%%%%%%%%%%

The authors thank the Editor, an Associate Editor, and an anonymous referee for providing helpful suggestions that improved this paper. This research was partially supported by 4-VA, a collaborative partnership for advancing the Commonwealth of Virginia, by National Science Foundation under Grant CNS-1650540 to Virginia Tech, and by the Virginia Tech College of Science Research Equipment Fund. The authors acknowledge the Advanced Research Computing program at Virginia Tech for providing computational resources. This study would not have been possible without the prior efforts of those who gathered specimens from the field and deposited them in accessible museum collections. Specimen loans to DSM were made possible by R. Brown, D. Kizirian, T. Nguyen, T. T. Nguyen, A. Ohler, A. Resetar, K. Thirakhupt, and J. Vindum.
We also would like to thank the American Museum of Natural History, California Academy of Science, Chulalongkorn University Museum of Zoology, Field Museum of Natural History, Institute of Ecology and Biological Resources, Mus\'{e}um National d'Histoire Naturelle, University of Kansas Biodiversity Institute, North Carolina State Museum of Natural Sciences, and the Vietnam National Museum of Nature for providing specimens. DSM also would like to acknowledge the Department of Biology at James Madison University, where some samples were collected.

%%%%%%%%%%%%%%%%%%%%%%%%%%%%%%%%%%%%%%%%%%%%%%%%%%%%%%%%%%%%%%%%%%%%%%%%%%%%%%%%%%%%%%%%%%%%%%%%%
\section*{Conflicts of Interests/Competing Interests}
%%%%%%%%%%%%%%%%%%%%%%%%%%%%%%%%%%%%%%%%%%%%%%%%%%%%%%%%%%%%%%%%%%%%%%%%%%%%%%%%%%%%%%%%%%%%%%%%%
The authors have no conflicts of interests/competing interests.

	%%%%%%%%%%%%%%%%%%%%%%%%% ref %%%%%%%%%%%%%%%%%%%%%%%%%%%%%%%%%%%%%%%%%%%%%%%%%%%%%%%%%%%%%%%%%%%%%%%%%%%%%%%%%%%%%%%%%%%%%%%%%
	%\bibliographystyle{chicago}
	%\bibliography{ref}

\end{document}